\documentclass[11pt]{article}

\usepackage{colortbl}  
\usepackage{longtable}  

\usepackage[final]{acl}

\usepackage{times}
\usepackage{latexsym}
\usepackage{listings}
\usepackage{amsmath}

\usepackage[T1]{fontenc}

\usepackage[utf8]{inputenc}

\usepackage{microtype}

\usepackage{inconsolata}

\usepackage{graphicx}

\usepackage{times}
\usepackage{latexsym}
\usepackage{lipsum}
\usepackage[T1]{fontenc}

\usepackage[utf8]{inputenc}

\usepackage{microtype}

\usepackage{inconsolata}

\usepackage{graphicx}

\usepackage{hyperref}
\usepackage{url}
\usepackage{graphicx}
\usepackage{booktabs}
\usepackage{bm}
\usepackage{caption}
\usepackage{subcaption}
\usepackage{pifont}
\usepackage{multirow} 
\usepackage{xspace}
\usepackage{float}
\usepackage{siunitx}
\usepackage{placeins}
\usepackage{tabularx}

\usepackage{longtable}
\usepackage{array}
\usepackage{cleveref}

\usepackage[normalem]{ulem} 

\newcommand{\specialcell}[2][c]{\begin{tabular}[#1]{@{}c@{}}#2\end{tabular}}
\newcommand\eg[0]{\textit{e.g.}}

\title{Explorer: Scaling Exploration-driven Web Trajectory Synthesis for Multimodal Web Agents}

\author{
    Vardaan Pahuja$^1$$^{*\dagger}$,\quad
    Yadong Lu$^2$\footnotemark[1]$^{\P}$,\quad
    Corby Rosset$^2$,\quad
    Boyu Gou$^1$,\quad\\
    \textbf{Arindam Mitra$^2$,\quad
    Spencer Whitehead$^2$,\quad
    Yu Su$^1$,\quad
    Ahmed Awadallah$^2$} \\
    \textsuperscript{1}The Ohio State University\quad\quad
    \textsuperscript{2}Microsoft Research, Redmond \\
    {\tt pahuja.9@osu.edu, yadonglu@microsoft.com}
}

\newcommand{\model}{Explorer\xspace} 

\begin{document}
\maketitle
\renewcommand{\thefootnote}{\fnsymbol{footnote}}

\footnotetext[1]{Equal Contribution. $\dagger$ Work partly done during internship at Microsoft Research. $\P$ Project Lead.}

\renewcommand{\thefootnote}{\arabic{footnote}}

\begin{abstract}
Recent success in large multimodal models (LMMs) has sparked promising applications of agents capable of autonomously completing complex web tasks.
While open-source LMM agents have made significant advances in offline evaluation benchmarks, their performance still falls substantially short of human-level capabilities in more realistic online settings.
A key bottleneck is the lack of diverse and large-scale trajectory-level datasets across various domains, which are expensive to collect. 
In this paper, we address this challenge by developing a scalable recipe to synthesize the largest and most diverse trajectory-level dataset to date, containing over \num{94}K successful multimodal web trajectories, spanning \num{49}K unique URLs, \num{720}K screenshots, and \num{33}M web elements. 
In particular, we leverage extensive web exploration and refinement to obtain diverse task intents.
The average cost is \num{28} cents per successful trajectory, making it affordable to a wide range of users in the community.  
Leveraging this dataset, we train \textbf{\model}, a multimodal web agent, and demonstrate strong performance on both offline and online web agent benchmarks such as Mind2Web-Live, Multimodal-Mind2Web, and MiniWob++. 
Additionally, our experiments highlight data scaling as a key driver for improving web agent capabilities.
We hope this study makes state-of-the-art LMM-based agent research at a larger scale more accessible.\footnote{Project website: \url{https://osu-nlp-group.github.io/Explorer/}}
\end{abstract}

\section{Introduction}

Graphical User Interfaces (GUIs) serve as the primary medium for user interaction across digital environments.
Within the GUI environment, LLM-based agents \cite{language-agent-tutorial} have shown great potential in automating complex workflows for human users.
These agents are designed to operate across diverse interfaces, including the web \cite{mind2web, DBLP:conf/iclr/ZhouX0ZLSCOBF0N24, zheng2024gpt, zheng2025skillweaver}, desktop \cite{xie2024osworld, wu2024oscopilot}, and mobile platforms \cite{rawles2023androidinthewild, yan2023gpt}.
Navigating modern GUI interfaces, which integrate textual, graphical, and interactive components, typically requires agents to possess visual grounding, long-term planning, and memory management capabilities.


Recent work \cite{seeclick, gou2024uground} has demonstrated the effectiveness of synthetic data for enhancing visual grounding \cite{gou2024uground, chen2024guicourse, DBLP:conf/eccv/KapoorBRKKAS24, chen2024edge} and planning \cite{xu2024aguvis, zhang-etal-2024-android}.
Developing end-to-end GUI agents with long-term planning and grounding capabilities requires training on multi-step trajectory data \cite{xu2024agenttrek, xu2024aguvis, qin2025ui}.
However, existing trajectory datasets are primarily human-annotated \cite{mind2web, li2024on, DBLP:conf/icml/LuKR24} or leverage synthetic data just for task proposal curation \cite{DBLP:conf/kdd/LaiLIYCSYZZD024, chen2024guicourse}.
And human annotation is expensive to scale for collecting large and diverse training datasets.
Therefore, synthetic data has emerged as a promising alternative to human-annotated data~\cite{hartvigsen2022toxigen, sahu2022data, ye2022zerogen, tang2023does, mukherjee2023orca, mitra2024agentinstruct}.
Collecting trajectory-level datasets presents unique challenges:
1) curating a diverse set of task intents at scale,
2) deploying an agent capable of interacting with a real-world environment to complete these tasks through a series of actions, and
3) verifying whether the task is accomplished by the executed action sequence.

\begin{figure*}[htbp]
    \centering
    \includegraphics[width=\linewidth]{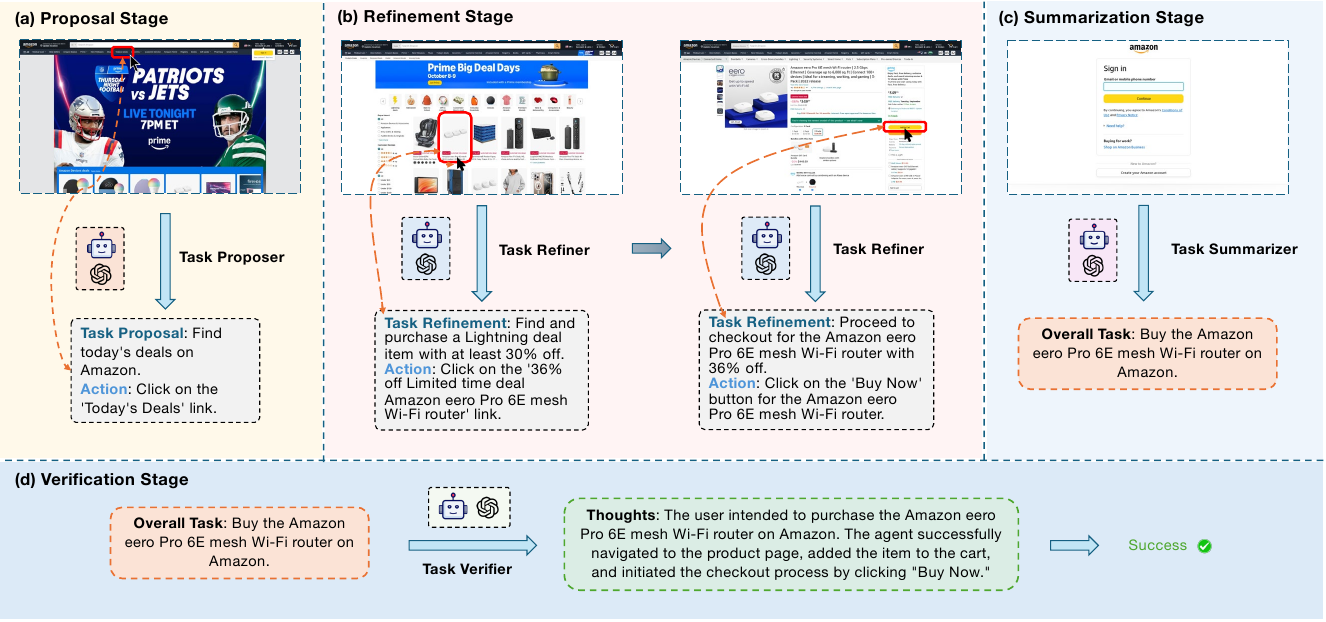}
    \caption{Data Generation Pipeline.
    The task proposer agent generates an abstract task proposal and the first action based on the website homepage.
    The task is then iteratively refined in subsequent steps by the refiner agent.
    Finally, the task summarizer agent constructs an overall task description from the action sequence, followed by task verification to assess correctness.
    }
    \label{fig:data_pipeline}
\end{figure*}


Data diversity is essential for equipping generalist web agents with a broad range of skills.
Existing work on synthetic web trajectory generation employs self-instruct for task proposal generation \cite{he2024openwebvoyager}.
It formulates task proposals from homepages or parametric LLM knowledge, overlooking the richer content available in deeper web pages, which is essential for achieving broader task diversity.
Another line of work leverages web tutorials as a form of supervision for generating web trajectories \cite{Ou2024SynatraTI, xu2024agenttrek}.
While web tutorials effectively cover common daily user tasks, the resulting trajectory data exhibits limited domain diversity in terms of website and domain coverage (Table~\ref{tab:data_comp}).
Additionally, information-seeking tasks remain underrepresented.
Due to these limitations, web agents trained on existing synthetic trajectory datasets have not seen much success in more realistic online evaluation settings.
To enhance web agents' performance in real-world settings, it is essential to incorporate greater diversity in their training trajectories.

\begin{table*}[htbp]
\centering
\small
\begin{tabular}{lccc} \toprule
      & \bfseries\# Trajectories & \# \bfseries Websites & \bfseries Modality               \\ \midrule
RUSS \cite{DBLP:conf/naacl/XuMDCHLL21} & \num{80} & \num{22} & HTML \\
Mind2Web \cite{mind2web} & \num{2350} & \num{137} & HTML + Screenshot \\
WebLINX \cite{DBLP:conf/icml/LuKR24} & \num{2337} & \num{155} & HTML + Screenshot \\
GUIAct \cite{chen2024guicourse} & \num{5696} & \num{121} & Screenshot\\
OpenWebVoyager \cite{he2024openwebvoyager} & \num{1165} & \num{48} & A11y tree + Screenshot \\
NNetnav \cite{murty2024nnetscape} & \num{6}K & \num{4} & A11y tree + Screenshot\\
AgentTrek \cite{xu2024agenttrek} & \num{10.4}K & \num{127} & A11y tree + HTML + Screenshot  \\
\midrule
\bfseries \model & \bfseries \num{94}K                & \bfseries \num{49}K            & \bfseries A11y tree + Screenshot (raw + SoM) + HTML \\ \bottomrule
\end{tabular}
\caption{Comparison to existing web agent benchmarks.
}
\label{tab:data_comp}
\end{table*}


In this work, we develop a \textit{scalable} and \textit{diverse} web trajectory data synthesis recipe for training GUI agent models.
Inspired by how humans learn to use the internet, \textit{we leverage exploration as a key mechanism for achieving diversity in task intents}.
We introduce \textbf{\model}, 
a framework for systematic web exploration to generate diverse, high-quality trajectory datasets.
Unlike prior work that relies on static task proposals, \model dynamically explores web environments to curate diverse, real-world tasks.
This exploration-based approach ensures broader task coverage and better generalization to real-world scenarios.
We instantiate this framework using popular URLs from several sources, such as Tranco \cite{DBLP:conf/ndss/PochatGTKJ19} and \url{similarweb.com} as seeds.
Our dataset comprises \num{94}K diverse web trajectories spanning \num{49}K unique URLs, making it the largest web trajectory dataset to date.
Each trajectory is richly annotated with artifacts such as screenshots, raw and set-of-mark \cite{yang2023set} annotated versions, HTML, and the accessibility tree, enabling comprehensive web agent training.
To construct this dataset, we develop a multi-agent pipeline that starts with an abstract task proposal and iteratively refines it into a more specific task through web exploration (Figure~\ref{fig:data_pipeline}).
Unlike previous approaches, our pipeline generates tasks better grounded in real-world websites, improving task relevance and diversity.
To demonstrate the effectiveness of our dataset, we train small language models using just the synthetic data and outperform existing web agent baselines by a significant margin.
The main contributions of this work are as follows:
\begin{itemize}

    \item We develop a scalable and easily customizable multi-agent pipeline for web agent trajectory synthesis. This pipeline leverages exploration as a core mechanism to generate diverse trajectory data, ensuring broad domain coverage and skill diversity in the resulting dataset.
    
    \item We leverage this pipeline to generate a diverse and high-quality GUI trajectory dataset consisting of \textbf{\num{94}K trajectories}, spanning \textbf{\num{49}K unique URLs} with \num{720}K screenshots and \num{33}M web elements, making it the largest web trajectory dataset of this scale.
    
   \item We demonstrate the effectiveness of our dataset by training small language models, which achieve strong performance on both online and offline benchmarks, significantly surpassing existing web agent baselines, including those with larger parameter counts.
\end{itemize}

\section{Related Work}

\begin{table*}
\small
\begin{tabular}{l}
\hline
\cellcolor[gray]{0.9}
\textbf{Information} \\
\hline
View the detailed 7-day weather forecast for Toronto, ON on The Weather Network website. \\
Convert 100 US Dollars to Euros using the XE currency converter. \\
Find directions from Seattle, WA to Bellevue, WA using Bing Maps. \\
\hline
\cellcolor[gray]{0.9}
\textbf{Service} \\
\hline
Research the French Bulldog breed on the American Kennel Club website, including its popularity and family life traits. \\
Find the nearest Penske truck rental location in Anaheim, California, and start the reservation process for a truck. \\
Explore and purchase a subscription for the UpToDate Pro Suite on the Wolters Kluwer website. \\
\hline
\cellcolor[gray]{0.9}
\textbf{Entertainment} \\
\hline
Find the Basscon presents: Darren Styles EDM event on Eventbrite, save it, and share it on Twitter. \\
View the details of the Photography Competition Winners - Season X and share the article on Twitter. \\
\hline
\cellcolor[gray]{0.9}
\textbf{Shopping} \\
\hline
Browse through the fall home decor section on the Target website to explore a variety of fall-themed home decor items. \\
Purchase a three-seat fabric sofa, specifically the UPPLAND Sofa, from IKEA's website. \\
\hline
\cellcolor[gray]{0.9}
\textbf{Travel} \\
\hline
Search for flights from Seattle to New York, select travel dates, and explore various flight options. \\
Find the weight of baggage allowance for economy class on qatarairways. \\
\hline
\end{tabular}
\caption{Example task descriptions from \model.}
\label{tab:traj_ex}
\end{table*}

Recent advances in multimodal language models have facilitated the development of web agents — autonomous systems designed to interact with real-world websites to perform everyday tasks \cite{mind2web, cogagent, seeclick, zheng2024gpt, zheng2025skillweaver, xue2025illusion}.
Early efforts to acquire trajectory data for training web agents primarily relied on crowd-sourcing \cite{mind2web, DBLP:conf/icml/LuKR24}.
However, due to the high cost of human annotation, recent work has adopted synthetic data generation for large-scale collection.
AutoWebGLM \cite{DBLP:conf/kdd/LaiLIYCSYZZD024} and GUIAct \cite{chen2024guicourse} utilize LLMs to generate task proposals, which human experts subsequently annotate.
OpenWebVoyager \cite{he2024openwebvoyager} employs a web agent to execute auto-generated task descriptions. 
However, since these task descriptions are generated using LLMs without exploring a website, they fail to capture the full diversity of possible tasks on that website.
Another line of work, including Synatra \cite{Ou2024SynatraTI} and AgentTrek \cite{xu2024agenttrek}, leverages web tutorials to guide web trajectory generation.
Meanwhile, concurrent effort \cite{murty2024nnetscape} employs an exploration-based trajectory generation in WebArena’s sandbox, while our work focuses on more realistic web agent evaluation on live websites.
To address diversity limitations in prior trajectory synthesis work, we design a bottom-up web trajectory synthesis pipeline that explores websites dynamically while maintaining a coherent high-level task intent.
We refer readers to Appendix~\ref{sec:related_appendix} for further discussion.


\section{Data Recipe}

\begin{table}[htbp]
\centering
\small
\begin{tabular}{@{}lc@{}}
\toprule
\bfseries Metric                        & \bfseries Value \\ \midrule
\# Total trajectories                         & \num{175}K  \\ 
\# Success trajectories                    & \num{94}K   \\ 
\# Unique URLs             & \num{49}K \\ 
Average steps per trajectory          & \num{7.7}   \\ 
Average elements per image   & \num{46.3}  \\ 
\midrule
\# Tokens           & \num{830}M  \\
\# Elements         & \num{33.3}M \\
\# Images           & \num{720}K\\ \midrule
Cost per trajectory         & \$\num{0.15} \\
Cost per successful trajectory         & \$\num{0.28} \\
\bottomrule
\end{tabular}
\caption{Dataset statistics for \model. The number of unique URLs, average steps per trajectory, average elements per image, and number of tokens, elements, and images correspond to the successful trajectories.
}
\label{tab:data_stat}
\end{table}

We design an automatic web trajectory synthesis pipeline that explores websites to generate diverse web trajectories.
It utilizes Playwright\footnote{https://playwright.dev/} to execute actions and collect metadata from real-world websites, starting from an initial URL.\footnote{For a \num{4}K subset of trajectories, we instruct GPT-4o to navigate to the target website by formulating a Google search query based on the task description.}
The metadata includes screenshots, HTML,  A11y tree, and actions in grounded and natural language forms.

\subsection{Website Selection}
We use a combination of URL sources to generate the synthetic web trajectories.
We obtain the top \num{100} URLs from \url{similarweb.com} corresponding to the high-traffic portion of the web with transactional tasks like booking flights, restaurant reservations, government services, sports, entertainment, etc.
The Tranco \cite{DBLP:conf/ndss/PochatGTKJ19} URLs include \num{49}K URLs representing the head portion of the web, which is less trafficked but popular nonetheless.
We filter out harmful websites containing violent or explicit content to ensure safety compliance.
Overall, we generate \num{94}K trajectories across both sources.
The complete data generation takes \num{50} hours, utilizing \num{60} parallel processes.
The viewport resolution is up to $1980\times1080$.







\subsection{Data Generation Pipeline}


We aim to develop a generalized pipeline for web exploration to collect diverse web trajectory data.
To enhance diversity, we adopt a bottom-up approach, starting with low-level actions and progressively shaping them into high-level task descriptions while maintaining a coherent task intent.
In the first step, the proposer agent generates an abstract task, which is refined to a more specific task through a refinement process (Figure~\ref{fig:data_pipeline}).
Since the agents execute actions alongside the refinement process, the generated tasks respect real-world constraints, such as product availability, available color options, and other specifications, ensuring practical applicability.
Our pipeline consists of the following LLM-powered agents\footnote{We use GPT-4o as the agent backbone throughout the data generation process.}:

\paragraph{Task Proposer.}
Given a website homepage, including its screenshot and accessibility tree, the task proposer agent generates diverse initial tasks that could be performed on that website.
The task descriptions at this stage are instructed to be high-level and abstract versions of the real-world tasks, which will be refined into more specific tasks in later stages.
Along with generating the task proposal, the agent proposes and executes the first action toward completing that task.
Furthermore, the agent is instructed to halt upon encountering robot detection such as CAPTCHA verification, login prompts, or payment requests.

\paragraph{Task Refiner.}
The task refiner agent receives the initial task proposal or the refined task description from the previous step, along with the corresponding action history as input.
It then predicts the next action consistent with the input task description and the updated, refined task description 
while incorporating the complete action history.
By iteratively refining the task description after each action, the agent ensures that the updated task remains aligned with the action history.

\begin{figure*}[htbp]
    \centering
    \includegraphics[width=0.7\linewidth]{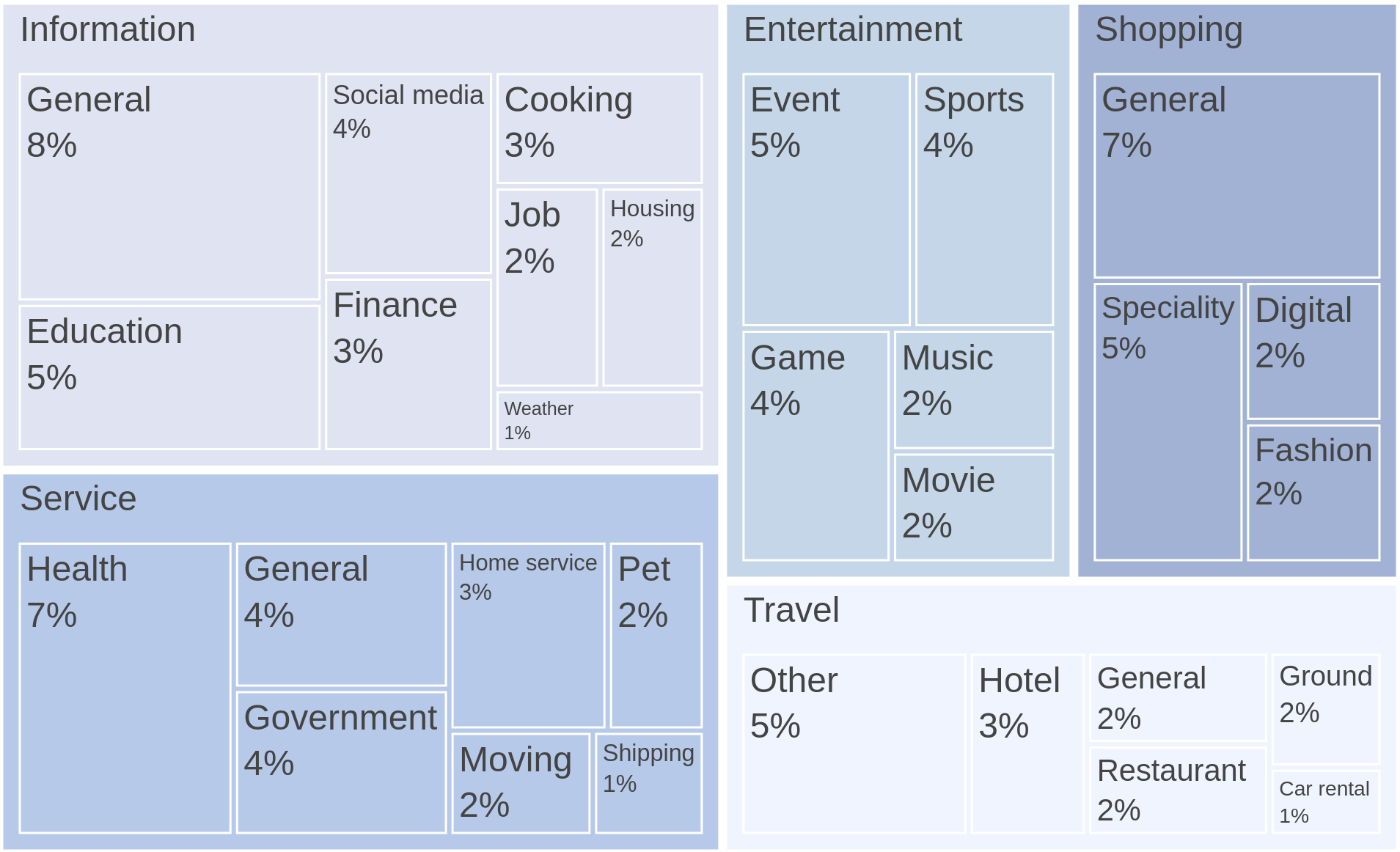}
    \caption{Data composition for \model.
    Its extensive diversity showcases its potential to train end-to-end generalist web agents.}
    \label{fig:data_composition}
\end{figure*}

\paragraph{Task Summarizer.}
This module processes the entire action and screenshot history to predict an overall task description that aligns with the trajectory.
The task summary is expected to be high level, \textit{i.e.}, it should describe what the task entails while omitting how it is accomplished.

\paragraph{Task Verifier.}\label{sec:verifier}
Inspired by \citet{DBLP:journals/corr/abs-2404-06474}, the task verifier agent receives the task description and action history, serving as a critic to evaluate whether the trajectory successfully completes the specified task.
In addition to the screenshots of the trajectory, it also receives a markdown representation of the last page.
This ensures the verifier has the full context of the website's final state, even when the viewport cannot capture all the content.
To ensure data quality, trajectories that are incoherent or misaligned with the high-level intent are discarded during this stage.
Such automatic evaluation of web trajectories has been widely adopted in prior work \cite{xu2024agenttrek, DBLP:conf/acl/HeYM0D0L024, koh2024tree}. 
Figure~\ref{fig:data_pipeline} illustrates the above pipeline.
The prompts for the above agents are given in Appendix~\ref{sec:prompt_details}. 





\subsection{Dataset Analysis}


\model comprises web trajectories spanning diverse domains, including services, entertainment, shopping, travel, and information, ensuring broad task diversity.
Sample tasks from \model are presented in Table~\ref{tab:traj_ex}.
Figure~\ref{fig:data_composition} visualizes the domain and subdomain distribution, highlighting the dataset's rich diversity.
To the best of our knowledge, \model with \num{94}K trajectories is the largest web trajectory dataset of this scale.
Table~\ref{tab:data_comp} shows a comparison with existing web agent datasets from the literature.
The detailed statistics are given in Table~\ref{tab:data_stat}. 
Beyond diversity, \model is also highly scalable and cost-efficient.
Our approach achieves a cost of \$\num{0.28} per successful trajectory, making it approximately $2\times$ more cost-effective than AgentTrek \cite{xu2024agenttrek} (which incurs \$\num{0.55} per trajectory) and significantly cheaper than human annotation (Table~\ref{tab:cost}).
Unlike human annotation, which requires training crowd workers and continuous quality monitoring, Explorer’s automated generation pipeline eliminates these bottlenecks, ensuring scalability with minimal overhead.
By combining diversity, scalability, and cost efficiency, \model sets a new benchmark for generating large-scale web trajectory datasets, making it an invaluable resource for training generalist GUI agents.

\begin{table}[htbp]
\centering
\small
\begin{tabular}{lc}
\toprule
\bfseries Model     & \bfseries Cost per trajectory\ \\ \midrule
Mind2Web \cite{mind2web} & \$\num{0.85} \\
AgentTrek \cite{xu2024agenttrek} & \$\num{0.55}     \\
\bfseries \model  & \$\num{0.28}    \\ 
\bottomrule
\end{tabular}
\caption{Cost comparison with other approaches.}
\label{tab:cost}
\end{table}


\section{Experiments}

\begin{table*}[htbp]
\centering
\small
\resizebox{\linewidth}{!}{%
\begin{tabular}{llllll}
\toprule
\bfseries Model                                                         & \bfseries Avg.\   Step SR (\%) & \bfseries Completion Rate  (\%) & \bfseries \specialcell{Task SR (1) (\%)} & \bfseries Full Task SR (\%)  \\ \midrule
\multicolumn{3}{l}{\textbf{API-based Models}} &\\
\cmidrule(r){1-1}
GPT-4o                                                                     & \num{58.5}                                          & \num{52.8}                                                                   & \num{44.6}  & \num{25.3}                              \\
GPT-3.5                                                                     & --                                          & \num{36.5}                                            & --       & \num{15.4}                                              \\ 
\midrule
\multicolumn{3}{l}{\textbf{Open-source Instructed Models}} &\\ \cmidrule(r){1-1}
Mistral-7B-Instr. \cite{jiang2023mistral}   &                       \num{32.8}                                      & \num{29.5}                                                                        & \num{24.1}        & \num{9.6}                        \\ 
Qwen2-72B-Instr. \cite{qwen} & -- & \textbf{\num{40.9}} & -- & \num{15.4} \\
Qwen2-VL-7B	\cite{Qwen2VL} & \num{40.2} &	\num{35.4}  &	\textbf{\num{34.9}}	&	\num{14.5} \\					
Phi-3.5V \cite{abdin2024phi}                                                                  & \num{28.5}                                      & \num{23.5}                                                                 & \num{20.5}   & \num{2.4}                          \\
\midrule
\multicolumn{3}{l}{\textbf{Supervised Fine-Tuning}} \\
\cmidrule(r){1-1}
\textbf{\model-4B} & \num{44.0} &	\num{39.4}	& \num{31.3} & \num{18.1}  \\ 
\textbf{\model-7B}   & \bfseries\num{45.3} &	\num{40.2}	 & \bfseries\num{34.9} & \textbf{\num{19.3}}\\
\bottomrule 
\end{tabular}
}
\caption{Results on Mind2Web-Live benchmark.
Missing values are denoted by --.
The results for GPT-4 and Mistral-7B-Instruct have been reproduced on our Linux servers.
The results for GPT-3.5 and Qwen2-72B-Instruct have been taken from \citet{pan2024webcanvas}.
The full task success rate represents the successful completion of all key nodes for a given task.
The average step success rate represents the proportion of completed key nodes, macro-averaged across tasks.
The completion rate represents the proportion of completed key nodes, micro-averaged across tasks.
Task SR (1) represents task SR with a tolerance of up to one error/key node.
}
\label{tab:m2w_live}
\end{table*}

We use the synthetic trajectories generated by our pipeline to train small multimodal language models (SLMs) for web agent tasks.
To ensure computational efficiency, we select \num{40}K trajectories from the full set for training.
We further refine this subset by filtering out trajectories that contain more than two scroll actions to mitigate potential model bias toward excessive scrolling behavior.
Finally, we use \textasciitilde\num{30}K trajectories obtained after filtering to fine-tune multimodal language models like Phi-3.5V \cite{abdin2024phi} and Qwen2-VL-7B \cite{Qwen2VL}.
For brevity, we denote the models trained on Phi-3.5V and Qwen2-VL-7B as \model-4B and \model-7B, respectively.
To test the effectiveness of our data  for web-based agentic tasks, we evaluate \model-4B and \model-7B
on Mind2Web-Live \cite{pan2024webcanvas}, Multimodal-Mind2Web \cite{mind2web,zheng2024gpt}, and MiniWob++ \cite{miniwob}.

\paragraph{Multimodal-Mind2Web.}
Multimodal-Mind2Web is an offline web agent benchmark comprising \num{2}K open-ended tasks spanning \num{137} websites across \num{31} domains. 
Each task comprises a sequence of actions with screenshots, action type, and HTML.
We follow the setting in \citet{zheng2024gpt}
and report element accuracy, operation F1, and step success rate (SR) as evaluation metrics.

\paragraph{Mind2Web-Live.}
Mind2Web-Live is a benchmark modified
from Mind2Web to test
web agents on live websites rather than static trajectories.
The benchmark evaluates performance using a key-node-based evaluation approach rather than using a golden action sequence, requiring valid trajectories to reach annotated ``key nodes'' across \num{104} test tasks in Mind2Web.
Since Mind2Web-Live relies on real-world dynamic websites, it encounters robot detection such as reCAPTCHA, which hinders testing \cite{xu2024aguvis}.
To address this, we select a subset of \num{83} test set tasks that remain consistently accessible throughout our tests.
Following \citet{pan2024webcanvas}, we report the average step success rate, completion rate, and full task success rate on the test set.

\paragraph{MiniWob++.} This benchmark consists of low-level tasks on a single webpage.
Typical examples include clicking a sequence of buttons, selecting items from a drop-down list, and filling out a form.
We use the subset of \num{46} tasks used for evaluation in prior work \cite{DBLP:conf/acl/ZengLLWLD024, Ou2024SynatraTI}.
The final score is obtained by averaging the results of four runs per task.
We use the zero-shot evaluation setting, which does not use any environment-specific trajectories for training.

\sisetup{detect-weight=true, detect-family=true}

\begin{table*}[htbp]

\centering
\small
\tabcolsep 3.5pt
\resizebox{\linewidth}{!}{%
\begin{tabular}{lllccccccccc|c}
\toprule
\multirow{2}{*}{} &
\multirow{2}{*}{\bfseries Model} &
\multirow{2}{*}{\bfseries Train Data} &
\multicolumn{3}{c}{\bfseries Cross-Task} &
\multicolumn{3}{c}{\bfseries Cross-Website} &
\multicolumn{3}{c}{\bfseries Cross-Domain} &
\multirow{2}{*}{\bfseries Avg.\ } \\
\cmidrule(r){4-6} \cmidrule(r){7-9} \cmidrule(r){10-12}
&  & & Ele. Acc & Op. F1 & Step SR & Ele. Acc & Op. F1 & Step SR & Ele. Acc & Op. F1 & Step SR \\
\midrule

&\multicolumn{8}{l}{\textbf{In-Context Learning}} &\\
\cmidrule(r){1-2}

&\textsc{GPT-3.5} & &\num{19.4}&\num{59.2}&\num{16.8}  
&\num{14.9}&\num{56.5}&\num{14.1}  
&\num{25.2}&\num{57.9}&\num{24.1}& \num{18.3}\\

&\textsc{GPT-4} & &\num{40.8}&\num{63.1}&\num{32.3}  
&\num{30.2}&\num{61.0}&\num{27.0}  
&\num{35.4}&\num{61.9}&\num{29.7}& \num{29.7}\\
&SeeAct \cite{zheng2024gpt} & &\num{46.4}&\num{73.4}&\num{40.2}  
&\num{38}&\num{67.8}&\num{32.4}  
&\num{42.4}&\num{69.3}&\num{36.8}& \num{36.5}\\

\midrule

&\multicolumn{8}{l}{\textbf{Supervised Fine-Tuning}} & &\\
\cmidrule(r){1-2}



& SeeClick-9.6B \cite{seeclick} & Syn.\ + M2W
& \num{26.3} & \num{86.2} & \num{23.7}
&\num{21.9} & \num{82.9} & \num{18.8}
& \num{22.1} & \num{84.1} & \num{20.2}& \num{20.9}\\

& EDGE-9.6B \cite{chen2024edge} & Syn.\ + M2W
& -- & -- & \num{30.0}
&-- & -- & \num{21.1} 
& -- & -- & \num{22.4} & \num{24.5}\\

& MiniCPM-3.1B \cite{chen2024guicourse} & Syn.\ + M2W & \num{23.8} & \num{86.8} & \num{20.8} & \num{20.3} & \num{81.7} & \num{17.3} & \num{17.9} & \num{74.5} & \num{14.6} & \num{17.6} \\

& ScribeAgent-32B \cite{shen2024scribeagent} & Syn.\ traj.\ & \num{38.0} & \num{52.9} & \num{35.6} & \num{34.1} & \num{52.7} & \num{32.5} & \num{39.4} & \num{54.7} & \num{37.3} & \num{35.1}\\

& AgentTrek-7B \cite{xu2024agenttrek} & Syn.\ + M2W  & \bfseries\num{60.8} & \num{88.9} & \bfseries\num{55.7} & \num{57.6} & \num{88.1} & \num{51.4} & \bfseries\num{56.0} & \num{87.5} & \num{52.6} & \num{53.2}\\
\midrule

& \bfseries \model-4B & Syn.\ traj.\ & \num{36.5}	& \num{82.9}	& \num{33.2}	& \num{44.1}	& \num{87.7}	& \num{39.3}	& \num{42.5}	& \num{86.3}	& \num{39.8} & \num{37.4}\\ 

& \bfseries \model-4B & M2W & \num{48.1}	& \num{88.0}	& \num{44.8}	& \num{49.1}	& \num{87.2}	& \num{45.0}	& \num{46.9}	& \num{87.7}	& \num{44.6} & \num{44.8}\\
& \bfseries \model-4B & Syn.\ + M2W & \num{53.4}	& \num{88.1}	& \num{50.7}	& \num{55.6}	& \num{89.5}	& \num{51.4}	& \num{49.8}	& \num{88.8}	& \num{47.2} & \num{49.8}\\ 
\midrule
& \bfseries \model-7B & Syn.\ traj.\ & \num{43.6}	& \num{86.6}	& \num{39.6}	& \num{48.7}	& \num{87.7}	& \num{44.5}	& \num{47.6}	& \num{87.2}	& \num{44.7} & \num{43.0}\\
& \bfseries \model-7B & M2W & \num{51.8}	& \num{88.0}	& \num{48.3}	& \num{56.3}	& \num{89.7}	& \num{52.0}	& \num{50.9}	& \num{88.9}	& \num{48.1} & \num{49.5}\\
 & \bfseries \model-7B & Syn.\ + M2W & \num{56.5}	& \bfseries\num{90.3}	& \num{53.2}	& \bfseries\num{60.5}	& \bfseries\num{90.7}	& \bfseries\num{56.7}	& \num{55.7}	& \bfseries\num{90.4}	& \bfseries\num{53.0} & \bfseries\num{54.3}\\

\bottomrule
\end{tabular}
}
\caption{Multimodal-Mind2Web evaluation results. The baseline numbers have been taken from \citet{zheng2024gpt, seeclick, chen2024edge, chen2024guicourse, shen2024scribeagent}. The last column denotes the average step success rates over the three test splits. \model significantly outperforms existing GUI agent baselines.
}
\label{tab:m2w_orig}
\end{table*}

\section{Results}


\subsection{In-domain Evaluation}
As an intrinsic evaluation of the trajectory collection pipeline, we generate \num{100} test tasks using \model, disjoint from the train set.
The SLM agents are tasked with executing the given tasks on live websites while an LLM-as-a-judge verifier (\S~\ref{sec:verifier}) evaluates the correctness of their actions at the trajectory level.
Table~\ref{tab:in_domain} shows the results.
We observe that the fine-tuned agents significantly outperform their pre-trained counterparts.
Thus, using in-domain web trajectory data training helps, which is a valuable sanity check.



\subsection{Mind2Web-Live Results} \label{sec:m2w_live_results}
We evaluate \model-4B and \model-7B trained on the synthetic trajectory dataset (Table~\ref{tab:m2w_live}).
We make the following observations from the results:\\

\paragraph{Improvement over base pre-trained models.}
We observe that \model-7B yields improvements of \num{5.1}\% and \num{4.8}\% in average step success rate (SR) and key node completion rate, respectively, compared to the pre-trained Qwen2-VL-7B model. 
Similarly, \model-4B obtains gains of \num{15.5}\% and \num{15.9}\% in average step SR and key node completion rate, respectively, over its pre-trained counterpart.
In terms of full task success rate, Phi-3.5V improves significantly from \num{2.4}\% to \num{18.1}\%, while Qwen2-VL-7B improves from \num{14.5}\% to \num{19.3}\%. 
To the best of our knowledge, this represents the state-of-the-art performance on Mind2Web-Live for models of this size trained exclusively on synthetic data.
\\

\begin{table}[htbp]
\centering
\small
\begin{tabular}{lc} \toprule
\bfseries Model             & \bfseries Full Task SR (\%) \\ \midrule
GPT-4o            & \num{16.0} \\ \midrule
Phi-3.5V & \num{1.0}                        \\
\bfseries\model-4B & \num{17.0}                        \\ \midrule
Qwen2-VL-7B & \num{6.0}                        \\
\bfseries\model-7B & \bfseries\num{18.0}                        \\ 
 \bottomrule                      
\end{tabular}
\caption{In-domain evaluation results. The fine-tuned \model models achieve significant improvements over their pre-trained counterparts and surpass closed-source LLMs, including GPT-4o.}
\label{tab:in_domain}
\end{table}

\paragraph{Improvement over higher capacity pre-trained models.}
Despite having much fewer parameters, we observe that \model-4B outperforms strong baselines such as Mistral-7B-Instruct-0.3 and Qwen2-72B-Instruct in full task SR by margins of \num{8.5}\% and \num{2.7}\%, respectively.
The Phi-3.5V model obtains an \num{18.1}\% full task success rate, which is better than GPT-3.5 (\num{15.4}\%), despite using orders of magnitude fewer parameters.
The corresponding results for the entire set of \num{104} tasks, including unreachable websites, are given in Appendix~\ref{sec:m2w_live_appendix}.
We provide the ablation studies in Appendix~\ref{sec:ablation} and the error analysis in Appendix~\ref{sec:m2w_live_err_analysis}.





\begin{figure*}[htbp]
    \begin{subfigure}[t]{\columnwidth}
    \centering
    \includegraphics[width=\linewidth]{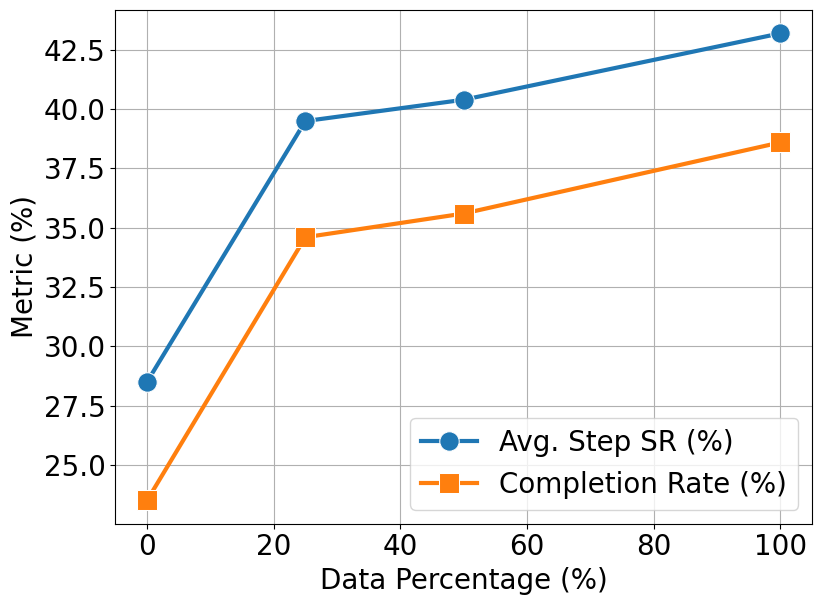}
    \end{subfigure}
    \hfill
    \begin{subfigure}[t]{\columnwidth}
    \centering
    \includegraphics[width=\linewidth]{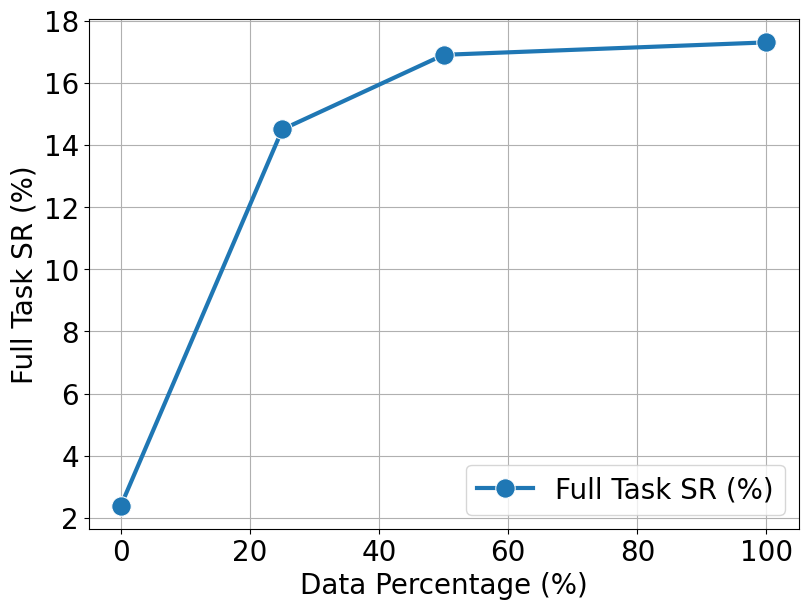}
\end{subfigure}
\caption{Experiments with data scaling using \model-4B on Mind2Web-Live. We experiment with using $100\%$, $50\%$, and $25\%$ of the trajectory data. All results are averaged over three runs. All metrics exhibit improvement with an increase in data scale.
}
\label{fig:data_scaling}
\end{figure*}

\subsection{Multimodal-Mind2Web Results} \label{sec:mm_m2w_results}
Following \citet{mind2web}, we obtain the top-\num{50} elements from a pre-trained DeBERTa \cite{he2021deberta} candidate generation model, which are then used to construct the accessibility tree and SoM image inputs.
The results are shown in Table~\ref{tab:m2w_orig}.

Among baselines, we include API-based models for in-context learning -- GPT-3.5, GPT-4, and SeeAct \cite{zheng2024gpt}. 
SeeAct is a web agent that performs web tasks using a two-step procedure of action generation and grounding using GPT-4V.
Additionally, we include baselines that fine-tune small language models using synthetic data, followed by further fine-tuning on the Mind2Web training set.
SeeClick \cite{seeclick} introduces a visual grounding model (Qwen-VL) trained on synthetically-generated grounding data.
EDGE \cite{chen2024edge} synthesizes QA data on webpages to improve the grounded GUI understanding capabilities of MLLMs.
ScribeAgent-Large \cite{shen2024scribeagent} and MiniCPM-GUI \cite{chen2024guicourse} use human-annotated trajectory data to train web agents.
AgentTrek \cite{xu2024agenttrek} is a GUI agent baseline that also utilizes synthetic trajectory data to fine-tune SLMs for Mind2Web, similar to our setting.
It synthesizes web trajectory data by guided replay from web tutorials.
We observe that \model-7B fine-tuned on synthetic data from \model plus Mind2Web outperforms all baselines in average step success rate.
Notably, it surpasses AgentTrek, which uses the same Qwen2-VL-7B MLLM backbone, highlighting the superior quality of our dataset.
The broad domain coverage and task diversity in \model contribute to its superior generalization across environments.


\begin{table}[htbp]
\centering
\small
\begin{tabular}{@{}ll@{}}
\toprule
\textbf{Model}       & \textbf{Accuracy (\%)} \\ \midrule
\multicolumn{2}{l}{\textit{API-based Models}} \\ \cmidrule(r){1-2}
GPT-3.5              & \num{39.57}                  \\
GPT-4                & \num{53.04}                  \\ \midrule
\multicolumn{2}{l}{\textit{Open-source Instructed Models}}\\ \cmidrule(r){1-2}
Phi-3.5V & \num{35.87} \\
Qwen2-VL-7B & \num{36.96} \\
Llama3-chat-8B       & \num{31.74}                  \\
Llama3-chat-70B      & \num{48.70}                  \\ \midrule
\multicolumn{2}{l}{\textit{Open-source Interactive Data Finetuned Models}}\\ \cmidrule(r){1-2}
AgentLM-7B \cite{DBLP:conf/acl/ZengLLWLD024}            & \num{15.65}                  \\
CodeActAgent-7B \cite{DBLP:conf/icml/WangCY0L0J24}       & \num{9.78}                   \\
AgentFlan-7B \cite{DBLP:conf/acl/ChenLWZLLCZ24}          & \num{20.87}                  \\
Lemur-chat-70B \cite{DBLP:conf/iclr/XuSXMLSHZLXCZKW24}        & \num{21.30}                  \\
AgentLM-70B \cite{DBLP:conf/acl/ZengLLWLD024}            & \num{36.52}                  \\
Synatra-CodeLlama-7B \cite{Ou2024SynatraTI}  & \num{38.20}                  \\
AgentTrek-7B \cite{xu2024agenttrek} & \num{45.28}\\
\midrule
\bfseries \model-4B                &  \num{46.74} \\
\bfseries \model-7B                & \bfseries \num{53.26} \\
\bottomrule                
\end{tabular}
\caption{Results on MiniWob++ benchmark \cite{miniwob} in zero-shot evaluation setting. The baseline numbers correspond to \citet{Ou2024SynatraTI}. \model outperforms much larger models by a significant margin.}
\label{tab:miniwob}
\end{table}

\begin{table}[htbp]
\centering
\small
\begin{tabular}{lr}
\toprule
\textbf{Statistic} & \textbf{Count} \\
\midrule
\# Unique task proposals     & \num{53}K \\
\# Unique final task descriptions & \num{81}K \\
\# Final task descriptions   & \num{94}K \\
\bottomrule
\end{tabular}
\caption{Task diversity statistics at different stages of the data synthesis pipeline.
Abstract task proposals evolve into diverse, fine-grained task descriptions as the agent explores the environment.}
\label{tab:task_stats}
\end{table}

\subsection{MiniWob++ Results}
Table~\ref{tab:miniwob} shows the results on the MiniWob++ benchmark in the zero-shot evaluation setting.
Among baselines, we have API-based models, in-context learning using open-source LMs, and agentic models like AgentLM \cite{DBLP:conf/acl/ZengLLWLD024}, CodeActAgent \cite{DBLP:conf/icml/WangCY0L0J24}, Lemur-Chat \cite{DBLP:conf/iclr/XuSXMLSHZLXCZKW24} and AgentFlan \cite{DBLP:conf/acl/ChenLWZLLCZ24} which include web-based demonstrations in their instruction tuning dataset.
Synatra-CodeLlama-7B \cite{Ou2024SynatraTI} and AgentTrek \cite{xu2024agenttrek} also synthesize web-agent trajectories automatically.
We observe that \model outperforms GPT-4 and general-purpose agent baselines.
\model-4B surpasses Synatra-CodeLlama-7B and AgentTrek-7B despite using a much smaller model with \num{4.2}B params, highlighting our synthetic data's superior quality and strong potential for generalization to new web environments.



\subsection{Data Scaling Experiments} \label{sec:data_scaling}
We conduct experiments with different data scales for \model-4B to analyze the impact of training data size.
Specifically, we subsample the original trajectory dataset to utilize \num{50}\% and \num{25}\% of its original size.
Figure~\ref{fig:data_scaling} presents the resulting performance curves.
Our results show that, even with just \num{25}\% of the training data, the model exhibits rapid performance gains over the base pre-trained model.
Increasing the dataset size further leads to gradual improvements across all reported metrics.
However, the increase in the overall task success rate is more gradual compared to the stepwise metrics, as it is a more coarse-grained metric.

\section{Analyses}

\paragraph{Diversity Analysis.}
We analyze the diversity of task descriptions across different stages of the data generation pipeline. At the initial proposal stage, the task proposer generates approximately \num{53}K unique high-level goals.
As the agent explores the web environment, these proposals evolve into a total of \num{94}K final task descriptions, of which \num{81}K are unique.
This progression shows that abstract task proposals can evolve into unique, fine-grained tasks with different constraints, shaped by the agent's specific exploration path.
Moreover, multiple trajectories corresponding to the same task description expose the agent to alternative solution paths, enhancing generalization.

\paragraph{Analysis of Verifier Accuracy.}
\begin{table}[htbp]
    \centering
    \small
    \begin{tabular}{lcc}
        \toprule
        & \textbf{Pred = Success} & \textbf{Pred = Failure} \\
        \midrule
        \textbf{GT = Success} & \textbf{\num{0.39}} & \num{0.05} \\
        \textbf{GT = Failure} & \num{0.14} & \textbf{\num{0.42}} \\
        \bottomrule
    \end{tabular}
    \caption{Confusion matrix of the task verifier evaluated on a subset of \num{100} Explorer-generated trajectories. We observe $81\%$ agreement with human judgment.}
    \label{tab:verifier_confusion}
\end{table}    
We did a human evaluation of the task verifier on a random set of \num{100} trajectories generated using Explorer. We obtain $81\%$ agreement with human judgement for the task verifier, which is comparable to prior work \cite{pan2024webcanvas, xu2024agenttrek}.

\begin{figure}
    \centering
    \includegraphics[width=\linewidth]{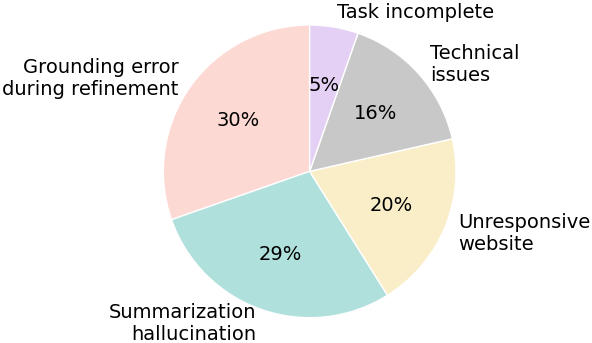}
    \caption{Error type distribution in synthetic trajectory synthesis. Grounding errors in the refinement phase and summarization hallucinations are the most dominant error types.}
    \label{fig:pipeline_error_types}
\end{figure}

\paragraph{Failure Modes of Trajectory Generation.}
To better understand failure cases in our pipeline, we perform a qualitative analysis of trajectories that were marked as unsuccessful by human annotators (Figure~\ref{fig:pipeline_error_types}).
The observed failure modes can be categorized as follows:
\begin{itemize}
    \item \textbf{Grounding error during refinement}: This error happens when the grounded action does not align with the natural language form of the agent's action during the task refinement phase.
    This misalignment propagates to the summarizer, resulting in an inaccurate final task description.
    
    \item \textbf{Unresponsive website}: The grounded action is correct and aligns with the task objective, but the website interaction fails due to non-responsive web elements during execution or dynamic content changes.

    \item \textbf{Summarization hallucination}:
    The summarizer introduces extraneous constraints or goals to the task description that are not present in the underlying trajectory, causing misalignment.

    \item \textbf{Technical issues}: The agent fails to complete the task due to external technical issues such as login requirements, media playback issues, or automated bot detection mechanisms.

    \item \textbf{Task incomplete}: The agent reaches the step limit before completing all required actions, resulting in an incomplete trajectory. 
\end{itemize}

\section{Conclusion}
In this work, we introduce \model, a scalable framework for synthesizing web trajectories on a large scale.
By leveraging thorough web exploration, \model ensures diversity in both domains and the skills acquired by web agents.
Unlike previous approaches, our framework generates contextually grounded trajectories that adapt to real-world constraints, improving both task relevance and generalization.
We instantiate this framework using URLs collected from diverse sources.
\model outperforms existing web agent baselines by a significant margin on both online and offline web agent benchmarks.
Furthermore, our results highlight the critical role of data scale in enhancing web agents' performance.
Future work will focus on extending this framework to encompass a broader range of GUI environments, such as operating systems with diverse applications.
GUI agents require specialized skills for different tasks, including information-seeking, operational, and navigation skills.
Efficient exploration of the environment to acquire these skills presents another promising avenue for future research.


\section*{Limitations}
\model explores the web environment autonomously, which may occasionally result in incoherent tasks.
Synthetic data collection using closed-source LLMs can be costly due to associated API expenses.
While this work serves as a proof of concept, future research will focus on developing tailor-made open-source LLMs for this task.
Additionally, some website content remains inaccessible due to login requirements, leading to insufficient data for those websites.

\section*{Ethical Considerations}
The synthetic data collection pipeline proposed in this paper is intended solely for academic research on GUI agents, with strict ethical safeguards to prevent unauthorized website interactions.
To ensure ethical compliance and mitigate risks, we prompt our agents to automatically terminate upon encountering CAPTCHA verifications, login prompts, or payment requests, ensuring that no actual transactions or bookings occur.
Additionally, we filter out websites containing violent or explicit content and strictly adhere to privacy regulations, ensuring that no personal information is used during action execution.
To enforce responsible data collection, we monitor a subset of automatically generated trajectories to ensure compliance with website access policies.
Moreover, we distribute the workload across websites to prevent excessive requests and minimize the impact on any single domain.

\section*{Acknowledgements}
We would like to thank colleagues from the OSU NLP group and Microsoft Research for their valuable feedback.
This research was supported in part by ARL W911NF2220144 and by computational resources provided by the Ohio Supercomputer Center \cite{center1987ohio}.

\FloatBarrier

\bibliography{custom}

\begin{thebibliography}{56}
\providecommand{\natexlab}[1]{#1}

\bibitem[{Abdin et~al.(2024)Abdin, Jacobs, Awan, Aneja, Awadallah, Awadalla, Bach, Bahree, Bakhtiari, Behl et~al.}]{abdin2024phi}
Marah Abdin, Sam~Ade Jacobs, Ammar~Ahmad Awan, Jyoti Aneja, Ahmed Awadallah, Hany Awadalla, Nguyen Bach, Amit Bahree, Arash Bakhtiari, Harkirat Behl, et~al. 2024.
\newblock Phi-3 technical report: A highly capable language model locally on your phone.
\newblock \emph{arXiv preprint arXiv:2404.14219}.

\bibitem[{Bai et~al.(2023)Bai, Bai, Chu, Cui, Dang, Deng, Fan, Ge, Han, Huang, Hui, Ji, Li, Lin, Lin, Liu, Liu, Lu, Lu, Ma, Men, Ren, Ren, Tan, Tan, Tu, Wang, Wang, Wang, Wu, Xu, Xu, Yang, Yang, Yang, Yang, Yao, Yu, Yuan, Yuan, Zhang, Zhang, Zhang, Zhang, Zhou, Zhou, Zhou, and Zhu}]{qwen}
Jinze Bai, Shuai Bai, Yunfei Chu, Zeyu Cui, Kai Dang, Xiaodong Deng, Yang Fan, Wenbin Ge, Yu~Han, Fei Huang, Binyuan Hui, Luo Ji, Mei Li, Junyang Lin, Runji Lin, Dayiheng Liu, Gao Liu, Chengqiang Lu, Keming Lu, Jianxin Ma, Rui Men, Xingzhang Ren, Xuancheng Ren, Chuanqi Tan, Sinan Tan, Jianhong Tu, Peng Wang, Shijie Wang, Wei Wang, Shengguang Wu, Benfeng Xu, Jin Xu, An~Yang, Hao Yang, Jian Yang, Shusheng Yang, Yang Yao, Bowen Yu, Hongyi Yuan, Zheng Yuan, Jianwei Zhang, Xingxuan Zhang, Yichang Zhang, Zhenru Zhang, Chang Zhou, Jingren Zhou, Xiaohuan Zhou, and Tianhang Zhu. 2023.
\newblock Qwen technical report.
\newblock \emph{arXiv preprint arXiv:2309.16609}.

\bibitem[{Center(1987)}]{center1987ohio}
Ohio~Supercomputer Center. 1987.
\newblock Ohio supercomputer center.

\bibitem[{Chen et~al.(2024{\natexlab{a}})Chen, Cui, Hu, Qin, Fang, Zhao, Wang, Liu, Chen, Huo et~al.}]{chen2024guicourse}
Wentong Chen, Junbo Cui, Jinyi Hu, Yujia Qin, Junjie Fang, Yue Zhao, Chongyi Wang, Jun Liu, Guirong Chen, Yupeng Huo, et~al. 2024{\natexlab{a}}.
\newblock Guicourse: From general vision language models to versatile gui agents.
\newblock \emph{arXiv preprint arXiv:2406.11317}.

\bibitem[{Chen et~al.(2024{\natexlab{b}})Chen, Li, Liang, Jiang, and Yang}]{chen2024edge}
Xuetian Chen, Hangcheng Li, Jiaqing Liang, Sihang Jiang, and Deqing Yang. 2024{\natexlab{b}}.
\newblock Edge: Enhanced grounded gui understanding with enriched multi-granularity synthetic data.
\newblock \emph{arXiv preprint arXiv:2410.19461}.

\bibitem[{Chen et~al.(2024{\natexlab{c}})Chen, Liu, Wang, Zhang, Liu, Lin, Chen, and Zhao}]{DBLP:conf/acl/ChenLWZLLCZ24}
Zehui Chen, Kuikun Liu, Qiuchen Wang, Wenwei Zhang, Jiangning Liu, Dahua Lin, Kai Chen, and Feng Zhao. 2024{\natexlab{c}}.
\newblock \href {https://doi.org/10.18653/v1/2024.findings-acl.557} {Agent-flan: Designing data and methods of effective agent tuning for large language models}.
\newblock In \emph{Findings of ACL}.

\bibitem[{Cheng et~al.(2024)Cheng, Sun, Chu, Xu, Li, Zhang, and Wu}]{seeclick}
Kanzhi Cheng, Qiushi Sun, Yougang Chu, Fangzhi Xu, Yantao Li, Jianbing Zhang, and Zhiyong Wu. 2024.
\newblock \href {https://doi.org/10.18653/v1/2024.acl-long.505} {Seeclick: Harnessing {GUI} grounding for advanced visual {GUI} agents}.
\newblock In \emph{Proceedings of ACL}.

\bibitem[{Deng et~al.(2023)Deng, Gu, Zheng, Chen, Stevens, Wang, Sun, and Su}]{mind2web}
Xiang Deng, Yu~Gu, Boyuan Zheng, Shijie Chen, Samual Stevens, Boshi Wang, Huan Sun, and Yu~Su. 2023.
\newblock \href {http://papers.nips.cc/paper\_files/paper/2023/hash/5950bf290a1570ea401bf98882128160-Abstract-Datasets\_and\_Benchmarks.html} {Mind2web: Towards a generalist agent for the web}.
\newblock In \emph{Proceedings of NeurIPS}.

\bibitem[{Gou et~al.(2025)Gou, Wang, Zheng, Xie, Chang, Shu, Sun, and Su}]{gou2024uground}
Boyu Gou, Ruohan Wang, Boyuan Zheng, Yanan Xie, Cheng Chang, Yiheng Shu, Huan Sun, and Yu~Su. 2025.
\newblock \href {https://openreview.net/forum?id=kxnoqaisCT} {Navigating the digital world as humans do: Universal visual grounding for {GUI} agents}.
\newblock In \emph{Proceedings of ICLR}.

\bibitem[{Gur et~al.(2024)Gur, Furuta, Huang, Safdari, Matsuo, Eck, and Faust}]{DBLP:conf/iclr/GurFHSMEF24}
Izzeddin Gur, Hiroki Furuta, Austin~V. Huang, Mustafa Safdari, Yutaka Matsuo, Douglas Eck, and Aleksandra Faust. 2024.
\newblock \href {https://openreview.net/forum?id=9JQtrumvg8} {A real-world webagent with planning, long context understanding, and program synthesis}.
\newblock In \emph{Proceedings of ICLR}.

\bibitem[{Hartvigsen et~al.(2022)Hartvigsen, Gabriel, Palangi, Sap, Ray, and Kamar}]{hartvigsen2022toxigen}
Thomas Hartvigsen, Saadia Gabriel, Hamid Palangi, Maarten Sap, Dipankar Ray, and Ece Kamar. 2022.
\newblock \href {https://doi.org/10.18653/v1/2022.acl-long.234} {Toxigen: {A} large-scale machine-generated dataset for adversarial and implicit hate speech detection}.
\newblock In \emph{Proceedings of ACL}.

\bibitem[{He et~al.(2024{\natexlab{a}})He, Yao, Ma, Yu, Dai, Zhang, Lan, and Yu}]{DBLP:conf/acl/HeYM0D0L024}
Hongliang He, Wenlin Yao, Kaixin Ma, Wenhao Yu, Yong Dai, Hongming Zhang, Zhenzhong Lan, and Dong Yu. 2024{\natexlab{a}}.
\newblock \href {https://doi.org/10.18653/v1/2024.acl-long.371} {Webvoyager: Building an end-to-end web agent with large multimodal models}.
\newblock In \emph{Proceedings of ACL}.

\bibitem[{He et~al.(2024{\natexlab{b}})He, Yao, Ma, Yu, Zhang, Fang, Lan, and Yu}]{he2024openwebvoyager}
Hongliang He, Wenlin Yao, Kaixin Ma, Wenhao Yu, Hongming Zhang, Tianqing Fang, Zhenzhong Lan, and Dong Yu. 2024{\natexlab{b}}.
\newblock Openwebvoyager: Building multimodal web agents via iterative real-world exploration, feedback and optimization.
\newblock \emph{arXiv preprint arXiv:2410.19609}.

\bibitem[{He et~al.(2021)He, Liu, Gao, and Chen}]{he2021deberta}
Pengcheng He, Xiaodong Liu, Jianfeng Gao, and Weizhu Chen. 2021.
\newblock \href {https://openreview.net/forum?id=XPZIaotutsD} {Deberta: decoding-enhanced bert with disentangled attention}.
\newblock In \emph{Proceedings of ICLR}.

\bibitem[{Hong et~al.(2024)Hong, Wang, Lv, Xu, Yu, Ji, Wang, Wang, Dong, Ding, and Tang}]{cogagent}
Wenyi Hong, Weihan Wang, Qingsong Lv, Jiazheng Xu, Wenmeng Yu, Junhui Ji, Yan Wang, Zihan Wang, Yuxiao Dong, Ming Ding, and Jie Tang. 2024.
\newblock \href {https://doi.org/10.1109/CVPR52733.2024.01354} {Cogagent: {A} visual language model for {GUI} agents}.
\newblock In \emph{Proceedings of CVPR}.

\bibitem[{Jiang et~al.(2023)Jiang, Sablayrolles, Mensch, Bamford, Chaplot, Casas, Bressand, Lengyel, Lample, Saulnier et~al.}]{jiang2023mistral}
Albert~Q Jiang, Alexandre Sablayrolles, Arthur Mensch, Chris Bamford, Devendra~Singh Chaplot, Diego de~las Casas, Florian Bressand, Gianna Lengyel, Guillaume Lample, Lucile Saulnier, et~al. 2023.
\newblock Mistral 7b.
\newblock \emph{arXiv preprint arXiv:2310.06825}.

\bibitem[{Kapoor et~al.(2024)Kapoor, Butala, Russak, Koh, Kamble, AlShikh, and Salakhutdinov}]{DBLP:conf/eccv/KapoorBRKKAS24}
Raghav Kapoor, Yash~Parag Butala, Melisa Russak, Jing~Yu Koh, Kiran Kamble, Waseem AlShikh, and Ruslan Salakhutdinov. 2024.
\newblock \href {https://doi.org/10.1007/978-3-031-73113-6\_10} {Omniact: {A} dataset and benchmark for enabling multimodal generalist autonomous agents for desktop and web}.
\newblock In \emph{Proceedings of ECCV}.

\bibitem[{Koh et~al.(2024)Koh, McAleer, Fried, and Salakhutdinov}]{koh2024tree}
Jing~Yu Koh, Stephen McAleer, Daniel Fried, and Ruslan Salakhutdinov. 2024.
\newblock Tree search for language model agents.
\newblock \emph{arXiv preprint arXiv:2407.01476}.

\bibitem[{Lai et~al.(2024)Lai, Liu, Iong, Yao, Chen, Shen, Yu, Zhang, Zhang, Dong, and Tang}]{DBLP:conf/kdd/LaiLIYCSYZZD024}
Hanyu Lai, Xiao Liu, Iat~Long Iong, Shuntian Yao, Yuxuan Chen, Pengbo Shen, Hao Yu, Hanchen Zhang, Xiaohan Zhang, Yuxiao Dong, and Jie Tang. 2024.
\newblock \href {https://doi.org/10.1145/3637528.3671620} {Autowebglm: {A} large language model-based web navigating agent}.
\newblock In \emph{Proceedings of KDD}.

\bibitem[{Li et~al.(2024)Li, Bishop, Li, Rawles, Campbell-Ajala, Tyamagundlu, and Riva}]{li2024on}
Wei Li, William~E Bishop, Alice Li, Christopher Rawles, Folawiyo Campbell-Ajala, Divya Tyamagundlu, and Oriana Riva. 2024.
\newblock \href {https://openreview.net/forum?id=yUEBXN3cvX} {On the effects of data scale on {UI} control agents}.
\newblock In \emph{Proceedings of NeurIPS Datasets and Benchmarks Track}.

\bibitem[{Liu et~al.(2018)Liu, Guu, Pasupat, Shi, and Liang}]{miniwob}
Evan~Zheran Liu, Kelvin Guu, Panupong Pasupat, Tianlin Shi, and Percy Liang. 2018.
\newblock \href {https://openreview.net/forum?id=ryTp3f-0-} {Reinforcement learning on web interfaces using workflow-guided exploration}.
\newblock In \emph{Proceedings of ICLR}.

\bibitem[{Liu et~al.(2023)Liu, Li, Wu, and Lee}]{DBLP:conf/nips/LiuLWL23a}
Haotian Liu, Chunyuan Li, Qingyang Wu, and Yong~Jae Lee. 2023.
\newblock \href {http://papers.nips.cc/paper\_files/paper/2023/hash/6dcf277ea32ce3288914faf369fe6de0-Abstract-Conference.html} {Visual instruction tuning}.
\newblock In \emph{Proceedings of NeurIPS}.

\bibitem[{Lu et~al.(2024)Lu, Kasner, and Reddy}]{DBLP:conf/icml/LuKR24}
Xing~Han Lu, Zdenek Kasner, and Siva Reddy. 2024.
\newblock \href {https://openreview.net/forum?id=mUSPhG4uDW} {Weblinx: Real-world website navigation with multi-turn dialogue}.
\newblock In \emph{Proceedings of ICML}.

\bibitem[{Mialon et~al.(2024)Mialon, Fourrier, Wolf, LeCun, and Scialom}]{mialon2024gaia}
Gr{\'e}goire Mialon, Cl{\'e}mentine Fourrier, Thomas Wolf, Yann LeCun, and Thomas Scialom. 2024.
\newblock \href {https://openreview.net/forum?id=fibxvahvs3} {{GAIA}: a benchmark for general {AI} assistants}.
\newblock In \emph{Proceedings of ICLR}.

\bibitem[{Mitra et~al.(2024)Mitra, Del~Corro, Zheng, Mahajan, Rouhana, Codas, Lu, Chen, Vrousgos, Rosset et~al.}]{mitra2024agentinstruct}
Arindam Mitra, Luciano Del~Corro, Guoqing Zheng, Shweti Mahajan, Dany Rouhana, Andres Codas, Yadong Lu, Wei-ge Chen, Olga Vrousgos, Corby Rosset, et~al. 2024.
\newblock Agentinstruct: Toward generative teaching with agentic flows.
\newblock \emph{arXiv preprint arXiv:2407.03502}.

\bibitem[{Mukherjee et~al.(2023)Mukherjee, Mitra, Jawahar, Agarwal, Palangi, and Awadallah}]{mukherjee2023orca}
Subhabrata Mukherjee, Arindam Mitra, Ganesh Jawahar, Sahaj Agarwal, Hamid Palangi, and Ahmed Awadallah. 2023.
\newblock Orca: Progressive learning from complex explanation traces of gpt-4.
\newblock \emph{arXiv preprint arXiv:2306.02707}.

\bibitem[{Murty et~al.(2024)Murty, Bahdanau, and Manning}]{murty2024nnetscape}
Shikhar Murty, Dzmitry Bahdanau, and Christopher~D Manning. 2024.
\newblock Nnetscape navigator: Complex demonstrations for web agents without a demonstrator.
\newblock \emph{arXiv preprint arXiv:2410.02907}.

\bibitem[{Ou et~al.(2024)Ou, Xu, Madaan, Liu, Lo, Sridhar, Sengupta, Roth, Neubig, and Zhou}]{Ou2024SynatraTI}
Tianyue Ou, Frank~F. Xu, Aman Madaan, Jiarui Liu, Robert Lo, Abishek Sridhar, Sudipta Sengupta, Dan Roth, Graham Neubig, and Shuyan Zhou. 2024.
\newblock \href {https://openreview.net/forum?id=KjNEzWRIqn} {Synatra: Turning indirect knowledge into direct demonstrations for digital agents at scale}.
\newblock In \emph{Proceedings of NeurIPS}.

\bibitem[{Pan et~al.(2024{\natexlab{a}})Pan, Zhang, Tomlin, Zhou, Levine, and Suhr}]{DBLP:journals/corr/abs-2404-06474}
Jiayi Pan, Yichi Zhang, Nicholas Tomlin, Yifei Zhou, Sergey Levine, and Alane Suhr. 2024{\natexlab{a}}.
\newblock \href {https://openreview.net/forum?id=NPAQ6FKSmK} {Autonomous evaluation and refinement of digital agents}.
\newblock In \emph{Proceedings of Conference on Language Modeling}.

\bibitem[{Pan et~al.(2024{\natexlab{b}})Pan, Kong, Zhou, Cui, Leng, Jiang, Liu, Shang, Zhou, Wu, and Wu}]{pan2024webcanvas}
Yichen Pan, Dehan Kong, Sida Zhou, Cheng Cui, Yifei Leng, Bing Jiang, Hangyu Liu, Yanyi Shang, Shuyan Zhou, Tongshuang Wu, and Zhengyang Wu. 2024{\natexlab{b}}.
\newblock \href {https://openreview.net/forum?id=O1FaGasJob} {Webcanvas: Benchmarking web agents in online environments}.
\newblock In \emph{Agentic Markets Workshop at ICML 2024}.

\bibitem[{Pochat et~al.(2019)Pochat, van Goethem, Tajalizadehkhoob, Korczynski, and Joosen}]{DBLP:conf/ndss/PochatGTKJ19}
Victor~Le Pochat, Tom van Goethem, Samaneh Tajalizadehkhoob, Maciej Korczynski, and Wouter Joosen. 2019.
\newblock \href {https://www.ndss-symposium.org/ndss-paper/tranco-a-research-oriented-top-sites-ranking-hardened-against-manipulation/} {Tranco: {A} research-oriented top sites ranking hardened against manipulation}.
\newblock In \emph{Proceedings of Network and Distributed System Security Symposium}.

\bibitem[{Qin et~al.(2025)Qin, Ye, Fang, Wang, Liang, Tian, Zhang, Li, Li, Huang et~al.}]{qin2025ui}
Yujia Qin, Yining Ye, Junjie Fang, Haoming Wang, Shihao Liang, Shizuo Tian, Junda Zhang, Jiahao Li, Yunxin Li, Shijue Huang, et~al. 2025.
\newblock Ui-tars: Pioneering automated gui interaction with native agents.
\newblock \emph{arXiv preprint arXiv:2501.12326}.

\bibitem[{Rawles et~al.(2023)Rawles, Li, Rodriguez, Riva, and Lillicrap}]{rawles2023androidinthewild}
Christopher Rawles, Alice Li, Daniel Rodriguez, Oriana Riva, and Timothy~P Lillicrap. 2023.
\newblock \href {https://openreview.net/forum?id=j4b3l5kOil} {Androidinthewild: A large-scale dataset for android device control}.
\newblock In \emph{Proceedings of NeurIPS Datasets and Benchmarks Track}.

\bibitem[{Sahu et~al.(2022)Sahu, Rodr{\'{\i}}guez, Laradji, Atighehchian, V{\'{a}}zquez, and Bahdanau}]{sahu2022data}
Gaurav Sahu, Pau Rodr{\'{\i}}guez, Issam~H. Laradji, Parmida Atighehchian, David V{\'{a}}zquez, and Dzmitry Bahdanau. 2022.
\newblock \href {https://doi.org/10.18653/v1/2022.nlp4convai-1.5} {Data augmentation for intent classification with off-the-shelf large language models}.
\newblock In \emph{Proceedings of the 4th Workshop on {NLP} for Conversational AI, ConvAI@ACL}.

\bibitem[{Shen et~al.(2024)Shen, Jain, Xiao, Amlekar, Hadji, Podolny, and Talwalkar}]{shen2024scribeagent}
Junhong Shen, Atishay Jain, Zedian Xiao, Ishan Amlekar, Mouad Hadji, Aaron Podolny, and Ameet Talwalkar. 2024.
\newblock Scribeagent: Towards specialized web agents using production-scale workflow data.
\newblock \emph{arXiv preprint arXiv:2411.15004}.

\bibitem[{Su et~al.(2024)Su, Yang, Yao, and Yu}]{language-agent-tutorial}
Yu~Su, Diyi Yang, Shunyu Yao, and Tao Yu. 2024.
\newblock \href {https://aclanthology.org/2024.emnlp-tutorials.3} {Language agents: Foundations, prospects, and risks}.
\newblock In \emph{Proceedings of EMNLP: Tutorial Abstracts}.

\bibitem[{Tang et~al.(2023)Tang, Han, Jiang, and Hu}]{tang2023does}
Ruixiang Tang, Xiaotian Han, Xiaoqian Jiang, and Xia Hu. 2023.
\newblock Does synthetic data generation of llms help clinical text mining?
\newblock \emph{arXiv preprint arXiv:2303.04360}.

\bibitem[{Wang et~al.(2024{\natexlab{a}})Wang, Bai, Tan, Wang, Fan, Bai, Chen, Liu, Wang, Ge, Fan, Dang, Du, Ren, Men, Liu, Zhou, Zhou, and Lin}]{Qwen2VL}
Peng Wang, Shuai Bai, Sinan Tan, Shijie Wang, Zhihao Fan, Jinze Bai, Keqin Chen, Xuejing Liu, Jialin Wang, Wenbin Ge, Yang Fan, Kai Dang, Mengfei Du, Xuancheng Ren, Rui Men, Dayiheng Liu, Chang Zhou, Jingren Zhou, and Junyang Lin. 2024{\natexlab{a}}.
\newblock Qwen2-vl: Enhancing vision-language model's perception of the world at any resolution.
\newblock \emph{arXiv preprint arXiv:2409.12191}.

\bibitem[{Wang et~al.(2024{\natexlab{b}})Wang, Chen, Yuan, Zhang, Li, Peng, and Ji}]{DBLP:conf/icml/WangCY0L0J24}
Xingyao Wang, Yangyi Chen, Lifan Yuan, Yizhe Zhang, Yunzhu Li, Hao Peng, and Heng Ji. 2024{\natexlab{b}}.
\newblock \href {https://openreview.net/forum?id=jJ9BoXAfFa} {Executable code actions elicit better {LLM} agents}.
\newblock In \emph{Proceedings of ICML}.

\bibitem[{Wu et~al.(2024)Wu, Han, Ding, Weng, Liu, Yao, Yu, and Kong}]{wu2024oscopilot}
Zhiyong Wu, Chengcheng Han, Zichen Ding, Zhenmin Weng, Zhoumianze Liu, Shunyu Yao, Tao Yu, and Lingpeng Kong. 2024.
\newblock \href {https://openreview.net/forum?id=3WWFrg8UjJ} {{OS}-copilot: Towards generalist computer agents with self-improvement}.
\newblock In \emph{Proceedings of ICLR 2024 Workshop on Large Language Model (LLM) Agents}.

\bibitem[{Xie et~al.(2024)Xie, Zhang, Chen, Li, Zhao, Cao, Hua, Cheng, Shin, Lei, Liu, Xu, Zhou, Savarese, Xiong, Zhong, and Yu}]{xie2024osworld}
Tianbao Xie, Danyang Zhang, Jixuan Chen, Xiaochuan Li, Siheng Zhao, Ruisheng Cao, Toh~Jing Hua, Zhoujun Cheng, Dongchan Shin, Fangyu Lei, Yitao Liu, Yiheng Xu, Shuyan Zhou, Silvio Savarese, Caiming Xiong, Victor Zhong, and Tao Yu. 2024.
\newblock \href {https://openreview.net/forum?id=tN61DTr4Ed} {{OSW}orld: Benchmarking multimodal agents for open-ended tasks in real computer environments}.
\newblock In \emph{Proceedings of NeurIPS Datasets and Benchmarks Track}.

\bibitem[{Xu et~al.(2021)Xu, Masling, Du, Campagna, Heck, Landay, and Lam}]{DBLP:conf/naacl/XuMDCHLL21}
Nancy Xu, Sam Masling, Michael Du, Giovanni Campagna, Larry Heck, James~A. Landay, and Monica Lam. 2021.
\newblock \href {https://doi.org/10.18653/v1/2021.naacl-main.80} {Grounding open-domain instructions to automate web support tasks}.
\newblock In \emph{Proceedings of NAACL}.

\bibitem[{Xu et~al.(2025{\natexlab{a}})Xu, Lu, Shen, Wang, Wang, Mao, Xiong, and Yu}]{xu2024agenttrek}
Yiheng Xu, Dunjie Lu, Zhennan Shen, Junli Wang, Zekun Wang, Yuchen Mao, Caiming Xiong, and Tao Yu. 2025{\natexlab{a}}.
\newblock \href {https://openreview.net/forum?id=EEgYUccwsV} {Agenttrek: Agent trajectory synthesis via guiding replay with web tutorials}.
\newblock In \emph{Proceedings of ICLR}.

\bibitem[{Xu et~al.(2024)Xu, Su, Xing, Mi, Liu, Shi, Hui, Zhou, Liu, Xie, Cheng, Zhao, Kong, Wang, Xiong, and Yu}]{DBLP:conf/iclr/XuSXMLSHZLXCZKW24}
Yiheng Xu, Hongjin Su, Chen Xing, Boyu Mi, Qian Liu, Weijia Shi, Binyuan Hui, Fan Zhou, Yitao Liu, Tianbao Xie, Zhoujun Cheng, Siheng Zhao, Lingpeng Kong, Bailin Wang, Caiming Xiong, and Tao Yu. 2024.
\newblock \href {https://openreview.net/forum?id=hNhwSmtXRh} {Lemur: Harmonizing natural language and code for language agents}.
\newblock In \emph{Proceedings of ICLR}.

\bibitem[{Xu et~al.(2025{\natexlab{b}})Xu, Wang, Wang, Lu, Xie, Saha, Sahoo, Yu, and Xiong}]{xu2024aguvis}
Yiheng Xu, Zekun Wang, Junli Wang, Dunjie Lu, Tianbao Xie, Amrita Saha, Doyen Sahoo, Tao Yu, and Caiming Xiong. 2025{\natexlab{b}}.
\newblock \href {https://arxiv.org/pdf/2412.04454} {Aguvis: Unified pure vision agents for autonomous gui interaction}.
\newblock In \emph{Proceedings of ICML}.

\bibitem[{Xue et~al.(2025)Xue, Qi, Shi, Song, Gou, Song, Sun, and Su}]{xue2025illusion}
Tianci Xue, Weijian Qi, Tianneng Shi, Chan~Hee Song, Boyu Gou, Dawn Song, Huan Sun, and Yu~Su. 2025.
\newblock An illusion of progress? assessing the current state of web agents.
\newblock \emph{arXiv preprint arXiv:2504.01382}.

\bibitem[{Yan et~al.(2023)Yan, Yang, Zhu, Lin, Li, Wang, Yang, Zhong, McAuley, Gao et~al.}]{yan2023gpt}
An~Yan, Zhengyuan Yang, Wanrong Zhu, Kevin Lin, Linjie Li, Jianfeng Wang, Jianwei Yang, Yiwu Zhong, Julian McAuley, Jianfeng Gao, et~al. 2023.
\newblock Gpt-4v in wonderland: Large multimodal models for zero-shot smartphone gui navigation.
\newblock \emph{arXiv preprint arXiv:2311.07562}.

\bibitem[{Yang et~al.(2023)Yang, Zhang, Li, Zou, Li, and Gao}]{yang2023set}
Jianwei Yang, Hao Zhang, Feng Li, Xueyan Zou, Chunyuan Li, and Jianfeng Gao. 2023.
\newblock Set-of-mark prompting unleashes extraordinary visual grounding in gpt-4v.
\newblock \emph{arXiv preprint arXiv:2310.11441}.

\bibitem[{Yao et~al.(2022)Yao, Chen, Yang, and Narasimhan}]{DBLP:conf/nips/Yao0YN22}
Shunyu Yao, Howard Chen, John Yang, and Karthik Narasimhan. 2022.
\newblock \href {http://papers.nips.cc/paper\_files/paper/2022/hash/82ad13ec01f9fe44c01cb91814fd7b8c-Abstract-Conference.html} {Webshop: Towards scalable real-world web interaction with grounded language agents}.
\newblock In \emph{Proceedings of NeurIPS}.

\bibitem[{Ye et~al.(2022)Ye, Gao, Li, Xu, Feng, Wu, Yu, and Kong}]{ye2022zerogen}
Jiacheng Ye, Jiahui Gao, Qintong Li, Hang Xu, Jiangtao Feng, Zhiyong Wu, Tao Yu, and Lingpeng Kong. 2022.
\newblock \href {https://doi.org/10.18653/v1/2022.emnlp-main.801} {Zerogen: Efficient zero-shot learning via dataset generation}.
\newblock In \emph{Proceedings of EMNLP}.

\bibitem[{Yoran et~al.(2024)Yoran, Amouyal, Malaviya, Bogin, Press, and Berant}]{DBLP:conf/emnlp/YoranAMBPB24}
Ori Yoran, Samuel~Joseph Amouyal, Chaitanya Malaviya, Ben Bogin, Ofir Press, and Jonathan Berant. 2024.
\newblock \href {https://aclanthology.org/2024.emnlp-main.505} {Assistantbench: Can web agents solve realistic and time-consuming tasks?}
\newblock In \emph{Proceedings of EMNLP}.

\bibitem[{Zeng et~al.(2024)Zeng, Liu, Lu, Wang, Liu, Dong, and Tang}]{DBLP:conf/acl/ZengLLWLD024}
Aohan Zeng, Mingdao Liu, Rui Lu, Bowen Wang, Xiao Liu, Yuxiao Dong, and Jie Tang. 2024.
\newblock \href {https://doi.org/10.18653/v1/2024.findings-acl.181} {Agenttuning: Enabling generalized agent abilities for llms}.
\newblock In \emph{Findings of ACL}.

\bibitem[{Zhang et~al.(2024)Zhang, Wu, Yihua, Liao, Xu, Xiao, Wei, and Tang}]{zhang-etal-2024-android}
Jiwen Zhang, Jihao Wu, Teng Yihua, Minghui Liao, Nuo Xu, Xiao Xiao, Zhongyu Wei, and Duyu Tang. 2024.
\newblock \href {https://aclanthology.org/2024.findings-emnlp.702/} {Android in the zoo: Chain-of-action-thought for {GUI} agents}.
\newblock In \emph{Findings of EMNLP}.

\bibitem[{Zheng et~al.(2025)Zheng, Fatemi, Jin, Wang, Gandhi, Song, Gu, Srinivasa, Liu, Neubig et~al.}]{zheng2025skillweaver}
Boyuan Zheng, Michael~Y Fatemi, Xiaolong Jin, Zora~Zhiruo Wang, Apurva Gandhi, Yueqi Song, Yu~Gu, Jayanth Srinivasa, Gaowen Liu, Graham Neubig, et~al. 2025.
\newblock Skillweaver: Web agents can self-improve by discovering and honing skills.
\newblock \emph{arXiv preprint arXiv:2504.07079}.

\bibitem[{Zheng et~al.(2024)Zheng, Gou, Kil, Sun, and Su}]{zheng2024gpt}
Boyuan Zheng, Boyu Gou, Jihyung Kil, Huan Sun, and Yu~Su. 2024.
\newblock \href {https://openreview.net/forum?id=piecKJ2DlB} {Gpt-4v(ision) is a generalist web agent, if grounded}.
\newblock In \emph{Proceedings of ICML}.

\bibitem[{Zhou et~al.(2024)Zhou, Xu, Zhu, Zhou, Lo, Sridhar, Cheng, Bisk, Fried, Alon et~al.}]{DBLP:conf/iclr/ZhouX0ZLSCOBF0N24}
Shuyan Zhou, Frank~F Xu, Hao Zhu, Xuhui Zhou, Robert Lo, Abishek Sridhar, Xianyi Cheng, Yonatan Bisk, Daniel Fried, Uri Alon, et~al. 2024.
\newblock \href {https://openreview.net/pdf?id=oKn9c6ytLx} {Webarena: A realistic web environment for building autonomous agents}.
\newblock In \emph{Proceedings of ICLR}.

\end{thebibliography}

\appendix


\label{sec:appendix}

\newpage

\setcounter{table}{0}
\renewcommand\thetable{\Alph{section}.\arabic{table}}
\setcounter{figure}{0}
\renewcommand\thefigure{\Alph{section}.\arabic{figure}}


\section*{Appendices}
In this supplementary material, we provide further details as follows:

\begin{itemize}
    \item Appendix~\ref{sec:m2w_eval}: Mind2Web Training and Evaluation Details
    \item Appendix~\ref{sec:cost_analysis}: Cost Analysis
    \item Appendix~\ref{sec:task_complexity}: Task Complexity Analysis
    \item Appendix~\ref{sec:prompt_details}: System Prompts
    \item Appendix~\ref{sec:traj_ex}: Trajectory Examples
    \item Appendix~\ref{sec:related_appendix}: More Related Work
\end{itemize}

\section{Mind2Web Training and Evaluation Details}\label{sec:m2w_eval}
Table~\ref{tab:hyper} shows the hyperparameters and training time for experiments on Mind2Web-Live and Multimodal-Mind2Web.
All experiments use Nvidia H100 GPUs.

\begin{figure*}[htbp]
    \centering
    \includegraphics[width=0.7\linewidth]{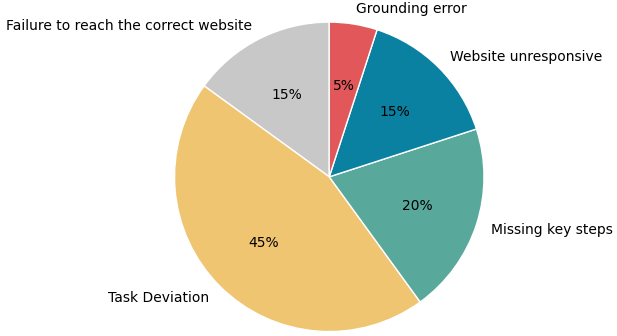}
    \caption{Statistics for different error cases in Mind2Web-Live evaluation. Task deviation is the most prevalent error type.
    }
    \label{fig:error_stats}
\end{figure*}

\begin{table*}[htbp]
\centering
\small
\resizebox{\linewidth}{!}{%
\begin{tabular}{llllll}
\toprule
\bfseries Model                                                          & \bfseries Avg.\   Step SR (\%) & \bfseries Completion Rate  (\%)  & \bfseries \specialcell{Task SR (1) (\%)} & \bfseries Full Task SR (\%)\\ \midrule
\multicolumn{3}{l}{\textbf{API-based Models}} &\\
\cmidrule(r){1-1}
GPT-4o                                                                     & \num{56.4}                                          & \num{50.4}                                               & \num{44.2}    & \num{22.1}                                                \\
GPT-3.5                                                                     & --                                          & \num{36.5}                                            & --  & \num{15.4}                                                   \\ \midrule
\multicolumn{3}{l}{\textbf{Open-source Instructed Models}} &\\ \cmidrule(r){1-1}

Mistral-7B-Instr. \cite{jiang2023mistral}                         & \num{33.0}                                      & \num{28.6}                                                & \num{25.0}    & \num{11.5}                                                    \\
Qwen2-72B-Instr. \cite{qwen} & -- & \bfseries\num{40.9} & -- & \num{15.4} \\
Qwen2-VL-7B \cite{Qwen2VL}	 &	\num{37.9} &	\num{33.3}  	& \num{31.7}	&	\num{12.5}\\
Phi-3.5V \cite{abdin2024phi}                                                                & \num{27.0}                                      & \num{22.3}                                           & \num{21.2}         & \num{1.9}                                          \\
\midrule
\multicolumn{3}{l}{\textbf{Supervised Fine-Tuning}} &\\
\cmidrule(r){1-1}
\textbf{\model-4B}  & \num{41.6}	& \num{36.7}	 & \num{30.8} & \num{16.4}\\
\textbf{\model-7B}  & \bfseries\num{42.0}	& \num{36.9}	 & \bfseries\num{32.7} & \bfseries\num{16.4}\\

\bottomrule
\end{tabular}
}
\caption{Results on Mind2Web-Live benchmark. 
The results for GPT-4, GPT-3.5, and Mistral-7B have been reproduced on our Linux servers.
The full task success rate (SR) represents the successful completion of all key nodes for a given task.
The average step success rate represents the proportion of completed key nodes, macro-averaged across tasks.
The completion rate represents the proportion of completed key nodes, micro-averaged across tasks.
Task SR (1) represents task SR with a tolerance of up to one error/key node.
Our Phi-3.5V model, finetuned on synthetic trajectory data from \model, outperforms much larger models, including Mistral-7B and Qwen2-72B-Instruct, by a significant margin and is comparable to GPT-3.5.
}
\label{tab:m2w_live_full}
\end{table*}

\subsection{Mind2Web-Live}\label{sec:m2w_live_appendix}
We exclude the following websites - \url{https://www.kbb.com}, \url{https://www.sixflags.com}, \url{https://www.viator.com}, \url{https://www.menards.com}, \url{https://www.amctheatres.com}, \url{https://www.cargurus.com}, \url{https://www.gamestop.com}, \url{https://www.cabelas.com}, \url{https://www.rei.com} due to denial of access faced during our tests.
Table~\ref{tab:m2w_live} shows the results on Mind2Web-Live for \num{83} out of \num{104} tasks across the remaining \num{37} websites.
The results on the whole Mind2Web-Live evaluation set are given in Table~\ref{tab:m2w_live_full}.
The results in Table~\ref{tab:m2w_live} are reported as the maximum over three runs, accounting for intermittent website access issues that may affect evaluation consistency.
For Mind2Web-Live, the dataloader first samples training instances at the trajectory level and then randomly samples a step from the trajectory to construct the final training instance.
Thus, the number of epochs is calculated at the trajectory level.
We use a viewport resolution of $1280\times 720$ during inference.
The Mind2Web-Live dataset is released under the MIT license, which permits its use in academic research.

\subsection{Multimodal-Mind2Web}
Following \citet{mind2web}, we obtain the top-\num{50} elements from a pre-trained DeBERTa \cite{he2021deberta} candidate generation model, which are then used to construct the accessibility tree and SoM image inputs.
Following \citet{Ou2024SynatraTI}, we always include the ground truth element in the input.
We use a viewport resolution of $1280\times 720$ which includes the GT element during inference.
We follow the setting in \citet{zheng2024gpt} and report element accuracy, operation F1, and step SR as evaluation metrics.
All experiments on Multimodal-Mind2Web use a single training and evaluation run.
The dataloader uniformly samples training instances from the set of action steps across all trajectories.
The Multimodal-Mind2Web dataset is released under the Responsible AI license, which permits its use in academic research.

\begin{table*}[htbp]
    \centering
    \small
    \resizebox{\linewidth}{!}{%
    \begin{tabular}{lcccc}
    \toprule
        \textbf{Dataset} & \textbf{Model} & \textbf{Train Data} & \textbf{Hyperparamerters} & \textbf{Train time (hours)}\\
        \midrule
        \multirow{3}{*}{M2W-Live} & Qwen2-VL-7B & Syn.\ &\texttt{batch\_size:\num{64}, epoch:\num{2}, learning\_rate:\num{1e-5}} & \num{15}\\
        & Qwen2-VL-7B & M2W &\texttt{batch\_size:\num{64}, epoch:\num{2}, learning\_rate:\num{1e-5}}& \num{1.5}\\
        & Qwen2-VL-7B & Syn.\ + M2W &\texttt{batch\_size:\num{64}, epoch:\num{2}, learning\_rate:\num{1e-5}}& \num{15.5}\\
        \midrule
        \multirow{3}{*}{M2W-Live} & Phi-3.5V & Syn.\ &\texttt{batch\_size:\num{64}, epoch:\num{2}, learning\_rate:\num{4e-5}}& \num{12.5}\\
        & Phi-3.5V & M2W &\texttt{batch\_size:\num{64}, epoch:\num{2}, learning\_rate:\num{1e-5}}& \num{1}\\
        & Phi-3.5V & Syn.\ + M2W &\texttt{batch\_size:\num{64}, epoch:\num{2}, learning\_rate:\num{4e-5}}& \num{12.5}\\
        \midrule
        \multirow{2}{*}{Multi.-M2W} & Qwen2-VL-7B & Syn.\ &\texttt{batch\_size:\num{64}, epoch:\num{10}, learning\_rate:\num{4e-5}}& \num{17}\\
        & Phi-3.5V & Syn.\ &\texttt{batch\_size:\num{64}, epoch:\num{10}, learning\_rate:\num{4e-5}}& \num{12}\\
    \bottomrule
    \end{tabular}
    }
    \caption{Hyperparameters used in our experiments.}
    \label{tab:hyper}
\end{table*}


\subsection{Ablation Studies}\label{sec:ablation}
We conduct ablation studies to assess the impact of various design choices on overall performance (Table~\ref{tab:m2w_live_diff_slm}).
To evaluate the importance of visual modality, we experiment with using just the textual modality for the Phi-3.5V model, replacing it with the text-only Phi-3-mini \cite{abdin2024phi}. In addition to Qwen2-VL-7B and Phi-3.5V, we also evaluate LLaVA-Mistral-7B \cite{DBLP:conf/nips/LiuLWL23a}, a strong MLLM baseline.
Our results show that omitting the visual modality leads to a sharp \num{4.8}\% drop in performance for Phi-3.5V, underscoring its importance for effective GUI grounding.
Furthermore, LLaVA-Mistral-7B significantly underperforms compared to both Qwen2-VL-7B and Phi-3.5V, highlighting the necessity of a stronger MLLM backbone for better GUI agent performance.

\begin{table*}[htbp]
\centering
\small
\begin{tabular}{llll}
\toprule
\bfseries Model                  & \bfseries Avg.\   Step SR (\%) & \bfseries Completion Rate  (\%) & \bfseries Full Task SR (\%) \\ \midrule
LLaVA-Mistral-7B       & \num{32.0}                                    & \num{30.3}                                      & \num{4.8}                              \\
Phi-3-mini (text-only) & \num{36.6}	& \num{34.0}	& \num{13.3} \\
Phi-3.5V               & \num{44.0}                                    & \num{39.4}                                     & \num{18.1}                              \\
Qwen2-VL-7B  & \bfseries\num{45.3} &	\bfseries\num{40.2}	& \textbf{\num{19.3}} \\
\bottomrule                            
\end{tabular}
\caption{Ablation studies on language models used for fine-tuning (Mind2Web-Live).
}
\label{tab:m2w_live_diff_slm}
\end{table*}


\subsection{Case Studies for Mind2Web-Live}\label{sec:m2w_live_err_analysis}

We randomly sample \num{20} error cases for \model on Mind2Web-Live to gain insights for future improvement.
These errors fall into the following categories:
\begin{itemize}
\item \textbf{Task deviation}: The agent executes actions unrelated to the given task, thus failing to complete it.
 
 
\item \textbf{Missing key steps}: The agent retrieves results that partially satisfy the required constraints, \eg, the agent finds women's clothes of the correct size but incorrect type or color.


\item \textbf{Grounding error}: The agent fails to interact with a valid element on the page.

\item \textbf{Website unresponsive}: The agent executes the correct action, but the website does not respond.

\item \textbf{Failure to reach the correct website}: This happens when the agent fails to output the correct website URL or use the search engine to arrive at the correct website.

\end{itemize}
Figure~\ref{fig:error_stats} presents the statistics for these error types.

\section{Cost Analysis}\label{sec:cost_analysis}

We use GPT-4o-turbo, which costs \$\num{2.5} per \num{1}M tokens for our trajectory synthesis.
Each proposal or refinement stage uses \num{3.6}K textual tokens on average.
Each input image costs \$\num{0.0028}.
The calculation assumes an average of \num{7.7} steps per trajectory, including the proposal stage.
Table~\ref{tab:cost_breakdown} shows the breakdown for the different stages of trajectory generation.
\begin{align*}
\textrm{Total cost} &= \$0.0128 * 7.7 + \$0.02581\\ &+ \$0.02381
= \$0.148
\end{align*}
The average cost per raw trajectory is \$\num{0.15}.
The success rate is estimated as \num{53.1}\%.
Thus, the average cost per successful trajectory is estimated to be \$\num{0.28}.

\begin{table}[H]
\centering
\small
\begin{tabular}{lll} \toprule
\bfseries Phase        & \bfseries Cost per step & \bfseries Total cost \\ \midrule
Proposal     & \$\num{0.0128}      & \$\num{0.0128}   \\
Refinement   & \$\num{0.0128}      & \$\num{0.0856}   \\
Verification & \$\num{0.02381}     & \$\num{0.02381}  \\
Summarization   & \$\num{0.02581}     & \$\num{0.02581} \\ \bottomrule
\end{tabular}
\caption{Cost breakdown for different modules in the pipeline.}
\label{tab:cost_breakdown}
\end{table}

\begin{table}[htbp]
\centering
\small
\begin{tabular}{@{}ll@{}}
\toprule
\bfseries Complexity level & \bfseries Count \\ \midrule
Easy             & \num{8.2}K  \\
Medium           & \num{44.3}K \\
Hard             & \num{41.2}K \\ \bottomrule
\end{tabular}
\caption{Task complexity statistics. Most tasks fall within the medium to high
complexity range.}
\label{tab:dataset_complexity}
\end{table}

\section{Task Complexity Analysis}\label{sec:task_complexity}

Explorer contains web trajectories spanning multiple steps with an average of $7.7$ steps per trajectory (Table~\ref{tab:data_stat}). Following prior work \cite{zheng2024gpt}, we use the number of action steps in a trajectory as a proxy for task complexity, categorizing tasks as \textit{easy} (2–4 steps), \textit{medium} (5–7 steps), and \textit{hard} (8–12 steps) in Table~\ref{tab:dataset_complexity}. 
We observe that the majority of tasks fall within the medium to high difficulty range, indicating a high level of complexity in the dataset.

\section{System Prompts}\label{sec:prompt_details}
The prompts for the task proposer agent, task refiner agent, task summarizer agent, task verifier agent, and captcha detection agent are given in Table~\ref{tab:proposal_prompt}, Table~\ref{tab:refiner_prompt}, Table~\ref{tab:summarizer_prompt}, Table~\ref{tab:verifier_prompt}, and Table~\ref{tab:captcha_prompt}, respectively.
The training prompt for \model is given in Table~\ref{tab:train_prompt}.

\onecolumn
{\small
\centering
\begin{longtable}{lp{12cm}}
    \hline
    \endfirsthead

    \multicolumn{2}{c}{\textit{Continued from previous page}} \\ \hline
    \endhead

    \hline \multicolumn{2}{|r|}{\textit{Continued on next page}} \\ \hline
    \endfoot

    \hline
    \endlastfoot

    \textbf{System Role} & What does this webpage show? Imagine you are a real user on this webpage. Given the webpage screenshot and parsed HTML/accessibility tree, please provide a single task that a user might perform on this page and the corresponding first action towards completing that task.\\
    & \underline{\smash{Do the following step by step}}:\\
    & 1. Generate a single task that a user might perform on this webpage. Be creative and come up with diverse tasks.\\
    & 2. Given the webpage screenshot and parsed HTML/accessibility tree, generate the first action towards completing that task (in natural language form).\\
    & 3. Given the webpage screenshot, parsed HTML/accessibility tree, and the natural language action, generate the grounded version of that action.\\~\\
    \cmidrule{2-2}

    & \textbf{ACTION SPACE}: Your action space is: [`click [element ID]', `type [element ID] [content]', `select [element ID] [content of option to select]', `scroll [up]', `scroll [down]', and `stop'].\\
    & \underline{\smash{Action output should follow the syntax as given below}}:\\
    & `click [element ID]': This action clicks on an element with a specific ID on the webpage.\\
    & `type [element ID] [content]': Use this to type the content into the field with id. By default, the "Enter" key is pressed after typing. Both the content and the ID should be within square braces as per the syntax. \\
    & `select [element ID] [content of option to select]': Select an option from a dropdown menu. The content of the option to select should be within square braces. When you get (select an option) tags from the accessibility tree, you need to select the serial number (element\textunderscore id) corresponding to the select tag, not the option, and select the most likely content corresponding to the option as input.\\
    & `scroll [down]': Scroll the page down. \\
    & `scroll [up]': Scroll the page up. \\~\\
    \cmidrule{2-2}
    
    & \textbf{IMPORTANT}: To be successful, it is important to STRICTLY follow the below rules:\\~\\
    & \textbf{Action generation rules}:\\
    & 1. You should generate a single atomic action at each step.\\
    & 2. The action should be an atomic action from the given vocabulary - click, type, select, scroll (up or down), or stop.\\
    & 3. The arguments to each action should be within square braces. For example, "click [127]", "type [43] [content to type]", "scroll [up]", "scroll [down]".\\
    & 4. The natural language form of action (corresponding to the field "action\textunderscore in\textunderscore natural\textunderscore language") should be consistent with the grounded version of the action (corresponding to the field "grounded \textunderscore action"). Do NOT add any additional information in the grounded action. For example, if a particular element ID is specified in the grounded action, a description of that element must be present in the natural language action. \\
    & 5. If the type action is selected, the natural language form of action ("action\textunderscore in\textunderscore natural\textunderscore language") should always specify the actual text to be typed. \\
    & 6. You should issue a “stop” action if the current webpage asks to log in or for credit card information. \\
    & 7. To input text, there is NO need to click the textbox first, directly type content. After typing, the system automatically hits the `ENTER' key.\\
    & 8. STRICTLY Avoid repeating the same action (click/type) if the webpage remains unchanged. You may have selected the wrong web element.\\
    & 9. Do NOT use quotation marks in the action generation.\\~\\
    \cmidrule{2-2}
    
    & \textbf{Task proposal rules}: \\
    & 1. You should propose tasks that are relevant to the website and can be completed using the website.\\
    & 2. You should only propose tasks that do not require login to execute the task.\\
    & 3. You should propose tasks that are clear and specific.\\
    & 4. For each task, provide concrete information or constraints, and use mock-up information (identifier, number, personal information, name, attributes, etc.) to make the task more specific and realistic.\\
    & 5. The task description should provide all the necessary information to complete the task.\\
    & 6. The task should be feasible to complete by a real user and should not require any additional information that is not available on the website.\\~\\

    & \underline{\smash{The output should be in below format}}:\\
    & \textbf{OUTPUT FORMAT}: Please give a short analysis of the screenshot, parsed HTML/accessibility tree, then put your answer within \textasciigrave\textasciigrave\textasciigrave \; \textasciigrave\textasciigrave\textasciigrave, for example, "In summary, the proposed task and the corresponding action is: \textasciigrave\textasciigrave\textasciigrave\texttt{{\{"task": <TASK>:str, "action\_in\_natural\_language":<ACTION\_IN\_NATURAL\_LANGUAGE>:str, "grounded\_action": <ACTION>:str\}}"}\textasciigrave\textasciigrave\textasciigrave\\
\midrule
\bfseries User Role & Website URL: \{\texttt{INIT\textunderscore URL}\}\\
& Parsed HTML\slash Accessibility Tree: \{\texttt{A11Y\textunderscore TREE}\}\\
& \{\texttt{SCREENSHOT}\} \\
\bottomrule        
\end{longtable}
}
\captionof{table}{Prompt for Task Proposer Agent.}
\label{tab:proposal_prompt}
\twocolumn

\nopagebreak
\onecolumn
{\small
\centering
\begin{longtable}{lp{12cm}}
    \hline
    \endfirsthead

    \multicolumn{2}{c}{\textit{Continued from previous page}} \\ \hline
    \endhead

    \hline \multicolumn{2}{|r|}{\textit{Continued on next page}} \\ \hline
    \endfoot

    \hline
    \endlastfoot

    \textbf{System Role} & What does this webpage show? Imagine you are a real user on this webpage, and your overall task is \{\texttt{OVERALL\textunderscore TASK}\}.
    This is the list of actions you have performed that lead to the current page \{\texttt{PREV\textunderscore ACTION\textunderscore LIST}\}. You are also given the webpage screenshot and parsed HTML/accessibility tree.\\
    & \underline{\smash{Do the following step by step}}:\\
    & 1. Please predict what action the user might perform next that is consistent with the previous action list in natural language.\\
    & 2. Then based on the parsed HTML/accessibility tree of the webpage and the natural language action, generate the grounded action.\\
    & 3. Update the overall task aligned with this set of actions.\\~\\

    \cmidrule{2-2}
    
    & \textbf{ACTION SPACE}: Your action space is: [`click [element ID]', `type [element ID] [content]', `select [element ID] [content of option to select]', `scroll [up]', `scroll [down]', and `stop'].\\
    & \underline{\smash{Action output should follow the syntax as given below}}:\\
    & `click [element ID]': This action clicks on an element with a specific id on the webpage.\\
    & `type [element ID] [content]': Use this to type the content into the field with id. By default, the "Enter" key is pressed after typing. Both the content and the id should be within square braces as per the syntax. \\
    & `select [element ID] [content of option to select]': Select an option from a dropdown menu. The content of the option to select should be within square braces. When you get (select an option) tags from the accessibility tree, you need to select the serial number (element\textunderscore id) corresponding to the select tag, not the option, and select the most likely content corresponding to the option as input.\\
    & `scroll [down]': Scroll the page down. \\
    & `scroll [up]': Scroll the page up. \\~\\

    \cmidrule{2-2}
    
    & \textbf{IMPORTANT}: To be successful, it is important to STRICTLY follow the below rules:\\~\\
    & \textbf{Action generation rules}:\\
    & 1. You should generate a single atomic action at each step.\\
    & 2. The action should be an atomic action from the given vocabulary - click, type, select, scroll (up or down), or stop.\\
    & 3. The arguments to each action should be within square braces. For example, "click [127]", "type [43] [content to type]", "scroll [up]", "scroll [down]".\\
    & 4. The natural language form of action (corresponding to the field "action\textunderscore in\textunderscore natural\textunderscore language") should be consistent with the grounded version of the action (corresponding to the field "grounded \textunderscore action"). Do NOT add any additional information in the grounded action. For example, if a particular element ID is specified in the grounded action, a description of that element must be present in the natural language action. \\
    & 5. If the type action is selected, the natural language form of action ("action\textunderscore in\textunderscore natural\textunderscore language") should always specify the actual text to be typed. \\
    & 6. You should issue the “stop” action when the given list of input actions is sufficient for a web task. \\
    & 7. You should issue a “stop” action if the current webpage asks to log in or for credit card information. \\
    & 8. To input text, there is NO need to click the textbox first, directly type content. After typing, the system automatically hits the `ENTER' key.\\
    & 9. STRICTLY Avoid repeating the same action (click/type) if the webpage remains unchanged. You may have selected the wrong web element.\\
    & 10. Do NOT use quotation marks in the action generation.\\~\\

    \cmidrule{2-2}

    & \textbf{Task proposal rules}:\\
    & 1. You should propose tasks that are relevant to the website and can be completed using the website itself.\\
    & 2. The overall task should be well-aligned to the entire set of actions in history plus the current generated action. It should not be focused just on the current action.\\
    & 3. You should only propose tasks that do not require login to execute the task.\\
    & 4. You should propose tasks that are clear and specific.\\
    & 5. For each task, provide concrete information or constraints, and use mock-up information (identifier, number, personal information, name, attributes, etc.) to make the task more specific and realistic.\\
    & 6. The task description should provide all the necessary information to complete the task.\\
    & 7. The task should be feasible to complete by a real user and should not require any additional information that is not available on the website.\\~\\

    & \underline{\smash{The output should be in below format}}:\\
    & \textbf{OUTPUT FORMAT}: Please give a short analysis of the screenshot, parsed HTML/accessibility tree, and history, then put your answer within \textasciigrave\textasciigrave\textasciigrave \; \textasciigrave\textasciigrave\textasciigrave, for example, "In summary, the proposed task and the corresponding action is: \textasciigrave\textasciigrave\textasciigrave\texttt{{\{"task": <TASK>:str, "action\_in\_natural\_language":<ACTION\_IN\_NATURAL\_LANGUAGE>:str, "grounded\_action": <ACTION>:str\}}"}\textasciigrave\textasciigrave\textasciigrave\\
    \midrule
\bfseries User Role & Website URL: \{\texttt{INIT\textunderscore URL}\}\\
& Parsed HTML\slash Accessibility Tree: \{\texttt{A11Y\textunderscore TREE}\}\\
& \{\texttt{SCREENSHOT}\} \\
\bottomrule        
\end{longtable}
}
\captionof{table}{Prompt for Task Refiner Agent.}
\label{tab:refiner_prompt}
\twocolumn

\nopagebreak
\begin{table*}[htbp]
    \centering
    \small
    \begin{tabular}{lp{12cm}}
    \toprule
        \textbf{System Role} & Given a list of actions performed on the website \{\texttt{WEBSITE\textunderscore URL}\} and the corresponding screenshots\\
        & List of actions: \{\texttt{ACTION\textunderscore LIST}\}\\
        & Your task is to come up with a single task description that will be accomplished by performing these actions in the given sequence on the website. \\~\\

    \cmidrule{2-2}
& \textbf{IMPORTANT}:\\
& 1. The task must contain some actions: ``Buy, Book, Find, Check, Choose, show me, search, browse, get, compare, view, give me, add to cart, ...'', ideally involving transactions/finding information on a specific product or service.\\
& 2. You should propose tasks that are clear and specific.\\
& 3. The task description should provide all the necessary information to complete the task.\\
& 4. The task description must indicate the domain of the website at the end of the task with the format: ``... on task website'', for instance, ``Purchase a laptop on Amazon'', ``Book a hair appointment on Yelp'', etc.\\
& 5. The task should be feasible to complete by a real user and should not require any additional information that is not specified in this input.\\
& 6. The task description should specify constraints like given budget, product features, and other specifications that can narrow down the search to a particular item/product.\\
& 7. Do NOT use any quotation marks (either single or double) in the task description.\\~\\

\cmidrule{2-2}

& \underline{\smash{The output should be in the below format}}:\\
& \textbf{OUTPUT FORMAT}: Please first give some analysis of the actions and screenshots and then output the overall task description. put your answer within \textasciigrave\textasciigrave\textasciigrave \; \textasciigrave\textasciigrave\textasciigrave, for example, ``In summary, the answer is: \textasciigrave\textasciigrave\textasciigrave\texttt{<TASK\textunderscore DESCRIPTION>:str}\textasciigrave\textasciigrave\textasciigrave''.\\
\midrule
\textbf{User Role} & \{\texttt{SCREENSHOT\textunderscore LIST}\}\\
    \bottomrule        
    \end{tabular}
    \caption{Prompt for Task Summarizer Agent.}
    \label{tab:summarizer_prompt}
\end{table*}

\begin{table*}[htbp]
    \centering
    \small
    \begin{tabular}{lp{12cm}}
    \toprule
        \textbf{System Role} & You are an expert in evaluating the performance of a web navigation agent. The agent is designed to help a human user navigate a website to complete a task. Given the user's intent, the agent's action history, and the final state of the webpage, your goal is to decide whether the agent's execution is successful or not.\\
& There are four types of tasks:\\~\\
& 1. \textbf{Transaction}: The user wants to perform a transaction on the webpage, such as booking a ticket, ordering a product, etc. The bot should at least initiate the add-to-cart or checkout process. It is still a success if the bot has done actions of `add to cart' or checkout and encounters the login page.  If the bot fails to do so, the task is considered a failure.\\~\\
& 2. \textbf{Information seeking}: The user wants to obtain certain information from the webpage, such as information of a product, reviews, map info, comparison of map routes, etc. The bot's response must contain the information the user wants, or explicitly state that the information is not available. Otherwise, e.g. the bot encounters an exception and responds with the error content, the task is considered a failure. Besides, be careful about the sufficiency of the agent's actions. For example, when asked to list the top-searched items in a shop, the agent should order the items by the number of searches, and then return the top items. If the ordering action is missing, the task is likely to fail.\\~\\
& 3. \textbf{Site navigation}: The user wants to navigate to a specific page. Carefully examine the bot's action history and the final state of the webpage to determine whether the bot successfully completes the task. No need to consider the bot's response.\\~\\
& 4. \textbf{Content modification}: The user wants to modify the content of a webpage or configuration. Carefully examine the bot's action history and the final state of the webpage to determine whether the bot successfully completes the task. No need to consider the bot's response.\\~\\

\cmidrule{2-2}
& \textbf{IMPORTANT}\\
& - If a product has been added to the bag/cart in the action list but just the purchase is pending, it should be counted as a success.\\
& - If you see the checkout page for the product you want to purchase, it should be counted as a success.\\
& - Format your response into two lines as shown below:\\~\\
& \textbf{Thoughts}: <your thoughts and reasoning process>\\
& \textbf{Status}: "success" or "failure"\\
\midrule
\textbf{User Role} & User Intent: \{\texttt{TASK\textunderscore DESCRIPTION}\}\\
& Action History: \{\texttt{ACTION\textunderscore HISTORY}\}\\
& The content of the last webpage in markdown format is given below: \{\texttt{LAST\textunderscore PAGE\textunderscore MARKDOWN}\}\\
& The snapshots of all webpages corresponding to the actions are shown in the images: \{\texttt{SCREENSHOT\textunderscore LIST}\}\\
    \bottomrule        
    \end{tabular}
    \caption{Prompt for Task Verifier Agent (adapted from \citet{DBLP:journals/corr/abs-2404-06474}).}
    \label{tab:verifier_prompt}
\end{table*}

\begin{table*}[htbp]
    \centering
    \small
    \begin{tabular}{lp{12cm}}
    \toprule
        \textbf{System Role} & You are an expert in evaluating whether the given webpage screenshot contains a captcha or not.\\
        & Given the last snapshot of the web page, your goal is to decide whether the webpage contains a captcha or not.\\
& Output ``Yes'' if the given webpage shows a captcha, otherwise ``No''. \\

    \cmidrule{2-2}
& \textbf{IMPORTANT}:\\
& Format your response into a line as shown below:\\
& Answer: ``Yes'' or ``No''\\

\midrule
\textbf{User Role} & The screenshot of the web page is shown in the image.\\
& \{\texttt{SCREENSHOT}\}\\
    \bottomrule        
    \end{tabular}
    \caption{Prompt for Captcha Detection Agent.}
    \label{tab:captcha_prompt}
\end{table*}

\begin{table*}[htbp]
    \centering
    \small
    \begin{tabular}{lp{12cm}}
    \toprule
        \textbf{System Role} & You are an expert at completing instructions on webpage screens.\\
               & You will be presented with a screenshot image with some numeric tags.\\
               & If you decide to click somewhere, you should choose the numeric element index closest to the location you want to click. \\
               & You should decide the action to continue this instruction.
               You will be given the accessibility tree of the current screen in the format: \texttt{[element\_idx] [role] [alt text or button name]}.\\
               & Here are the available actions:\\
               & \texttt{\{"action": "goto", "action\_natural\_language": str, "value": \textless the URL to go to\textgreater\}}\\
               & \texttt{\{"action": "google\_search", "action\_natural\_language": str, "value": \textless search query for google\textgreater\}}\\
               & \texttt{\{"action": "click", "action\_natural\_language": str, "idx": \textless element\_idx\textgreater\}}\\
               & \texttt{\{"action": "type", "action\_natural\_language": str, "idx": \textless element\_idx\textgreater, "value": \textless the text to enter\textgreater\}}\\
               & \texttt{\{"action": "select", "action\_natural\_language": str, "idx": \textless element\_idx\textgreater, "value": \textless the option to select\textgreater\}}\\
               & \texttt{\{"action": "scroll [up]", "action\_natural\_language": str\}}\\
               & \texttt{\{"action": "scroll [down]", "action\_natural\_language": str\}}\\
               & Your final answer must be in the above format.\\
        \midrule
        \textbf{User Role} & The instruction is to \{\texttt{TASK\textunderscore DESCRIPTION}\}. \\
      & History actions: \{\texttt{PREVIOUS\textunderscore ACTION\textunderscore LIST}\}\\
      & Here is the screen information: \{\texttt{A11Y\textunderscore TREE}\}\\
      & Think about what you need to do with the current screen, and output the action in the required format in the end. \\
        \bottomrule
    \end{tabular}
    \caption{Prompt for web agent training.}
    \label{tab:train_prompt}
\end{table*}

\FloatBarrier

\section{Trajectory Examples}\label{sec:traj_ex}
\begin{figure*}[htbp]
    \centering
    \includegraphics[width=\textwidth]{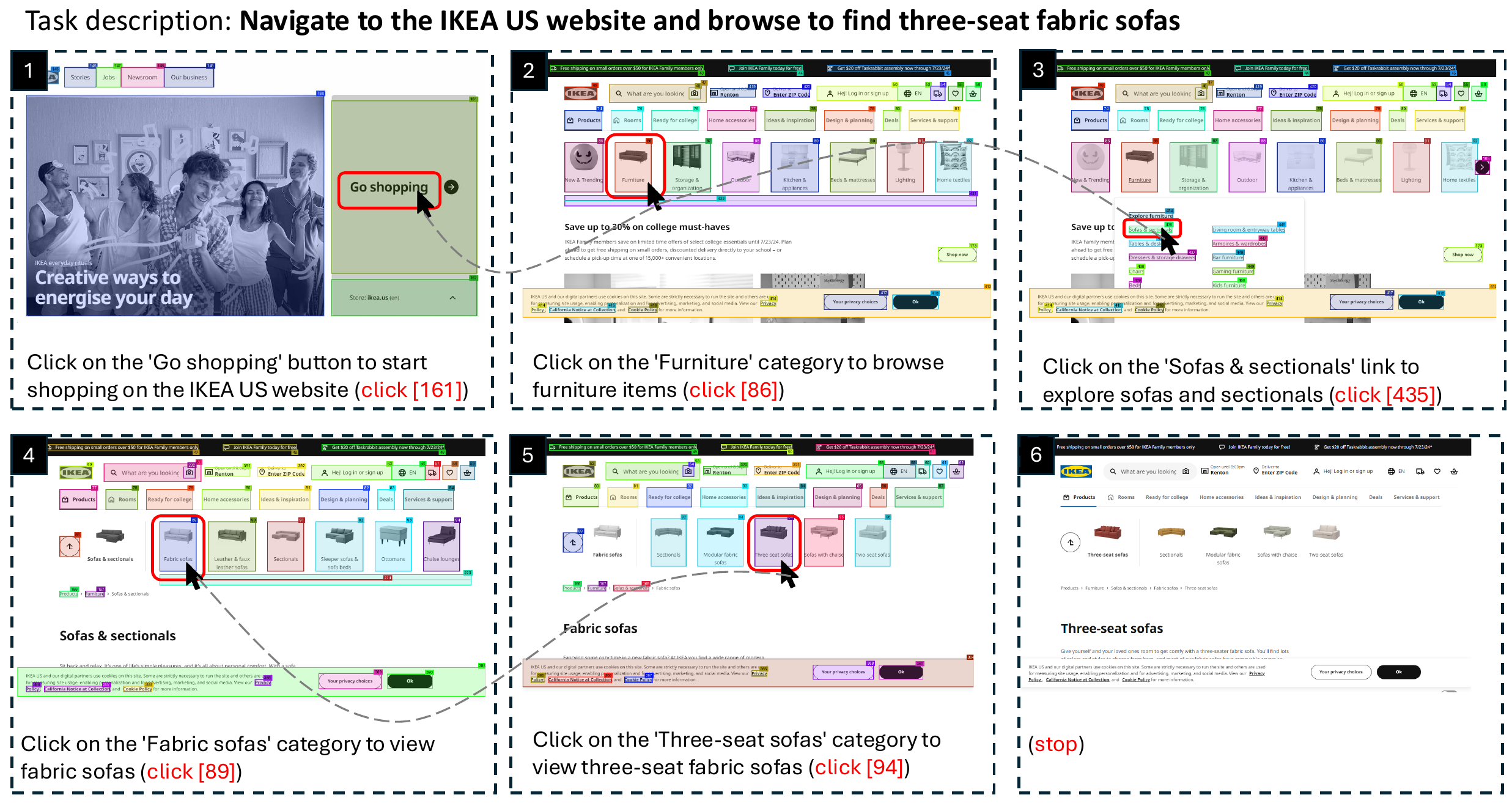}
    \caption{Example synthetic trajectory from \model. Each step shows the set-of-mark annotated screenshot along with the grounded action taken by the GPT-4 agent.}
    \label{fig:traj_ex}
\end{figure*}

\begin{figure*}[htbp]
    \centering
    \includegraphics[width=\textwidth]{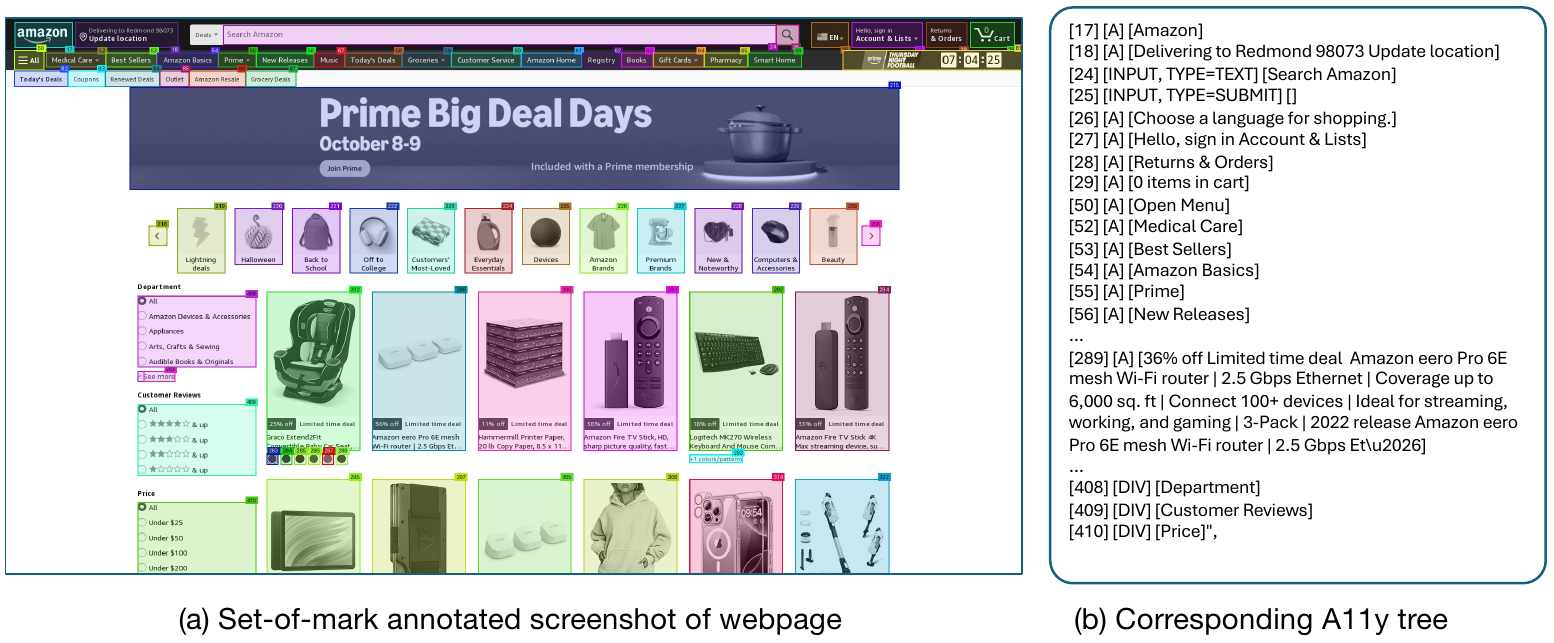}
    \caption{Visualization of the model inputs during trajectory generation, model training, and inference. The example corresponds to step 2 of the trajectory in Figure~\ref{fig:data_pipeline}.}
    \label{fig:traj_input}
\end{figure*}

Figure~\ref{fig:traj_ex} shows a sample trajectory executed on the IKEA website.
Figure~\ref{fig:traj_input} shows the set-of-mark annotations and accessibility tree inputs of the model during trajectory generation, training, and inference.
The action space for our trajectory synthesis pipeline is given in Table~\ref{tab:action}.


\begin{table*}[htbp]
\small
\centering
\begin{tabular}{llc}
\toprule
\textbf{Action Type}  & \textbf{Description} & \textbf{Count}\\
\midrule
click [$\mathtt{elem}$] & Click on $\mathtt{elem}$. & \num{415}K\\
type [$\mathtt{elem}$] [$\mathtt{text}$] & Type $\mathtt{text}$ & \num{62}K\\
select [$\mathtt{elem}$] [$\mathtt{text}$] & Select $\mathtt{text}$ from dropdown list. & \num{5}K\\
goto [$\mathtt{url}$] & Go to $\mathtt{url}$. & \num{26}K\\
search\textunderscore google [$\mathtt{query}$] & Search for $\mathtt{query}$ on Google. & \num{4}K\\
scroll [$\mathtt{up/down}$] & Scroll up or down. & \num{213}K \\
\bottomrule
\end{tabular}
\caption{Action space for web navigation in \model.
}
\label{tab:action}
\end{table*}

\section{More Related Work}\label{sec:related_appendix}
\subsection{LLM-based Web Agents}

Recent advances in multimodal language models have facilitated the development of web agents — autonomous systems designed to interact with real-world websites to perform everyday tasks \cite{mind2web, cogagent, seeclick, zheng2024gpt}.
Web agents have made significant progress, evolving from simulated environments \cite{miniwob} to complex real-world applications \cite{mind2web, DBLP:conf/nips/Yao0YN22, DBLP:conf/iclr/ZhouX0ZLSCOBF0N24}.
Key challenges for web agents include long-term planning, visual grounding, and memory management. 
To improve long-context understanding, WebAgent \cite{DBLP:conf/iclr/GurFHSMEF24} utilizes multiple LLMs - one for planning, summarization, and grounded program synthesis.
SeeAct \cite{zheng2024gpt} adopts a two-step procedure of planning followed by grounding at each step using GPT-4 to accomplish web agent tasks.
SkillWeaver \cite{zheng2025skillweaver} introduces a self-improving agent framework that autonomously synthesizes reusable skills as APIs through iterative exploration.
Another line of work employs a vision-only approach to train a GUI grounding model that directly predicts pixel coordinates for executing GUI agent tasks \cite{seeclick, DBLP:conf/eccv/KapoorBRKKAS24, gou2024uground}.
However, a significant bottleneck remains — the lack of large-scale, high-quality web trajectory data for training robust agents. 
Our work presents a new framework for synthesizing large-scale web trajectory data to train end-to-end web agents.



\subsection{Web Agent Benchmarks and Datasets}
Early benchmarks for web tasks such as MiniWob++ \cite{miniwob} focused on testing low-level actions on simulated websites. 
However, these simulated websites fail to capture the complexity of the real-world web.
Mind2Web \cite{mind2web} introduces a trajectory-level dataset with \num{2}K tasks across \num{137} real-world websites and \num{31} domains.
However, it employs a static evaluation method that penalizes alternative valid execution paths.
To overcome this limitation, follow-up work has explored alternative evaluation approaches, including functional correctness-based evaluation in WebArena \cite{DBLP:conf/iclr/ZhouX0ZLSCOBF0N24} and key-node-based evaluation in Mind2Web-Live \cite{pan2024webcanvas}.
Most recently, Online-Mind2Web \cite{xue2025illusion} addresses challenges in online evaluation, such as sensitivity to website updates, and demonstrates improved agreement with human judgments, providing a more reliable benchmark for real-world web agent evaluation.
Towards the goal of making web agents more capable of performing realistic tasks, GAIA \cite{mialon2024gaia} and AssistantBench \cite{DBLP:conf/emnlp/YoranAMBPB24} introduce benchmarks that include time-consuming information-seeking tasks.
In this work, we develop \model, a multimodal web agent trained on our synthetic dataset, and showcase its strong performance across online and offline benchmarks, including Mind2Web-Live, Multimodal-Mind2Web, and MiniWob++.

\end{document}